# Solving #S<span>AT</span> and Bayesian Inference with Backtracking Search

**Fahiem Bacchus**                                                    FBACCHUS@CS.TORONTO.EDU
**Shannon Dalmao**
**Toniann Pitassi**                                                        TONI@CS.TORONTO.EDU
*Department of Computer Science*
*University of Toronto*
*Toronto, Ontario*
*Canada, M5S 3G4*

## Abstract

Inference in Bayes Nets (BAYES) is an important problem with numerous applications in probabilistic reasoning. Counting the number of satisfying assignments of a propositional formula (#SAT) is a closely related problem of fundamental theoretical importance. Both these problems, and others, are members of the class of sum-of-products (SUMPROD) problems. In this paper we show that standard backtracking search when augmented with a simple memoization scheme (caching) can solve any sum-of-products problem with time complexity that is at least as good any other state-of-the-art exact algorithm, and that it can also achieve the best known time-space tradeoff. Furthermore, backtracking's ability to utilize more flexible variable orderings allows us to prove that it can achieve an exponential speedup over other standard algorithms for SUMPROD on some instances.

The ideas presented here have been utilized in a number of solvers that have been applied to various types of sum-of-product problems. These system's have exploited the fact that backtracking can naturally exploit more of the problem's structure to achieve improved performance on a range of problem instances. Empirical evidence of this performance gain has appeared in published works describing these solvers, and we provide references to these works.

## 1. Introduction

Probabilistic inference in Bayesian Networks (BAYES) is an important and well-studied problem with numerous practical applications in probabilistic reasoning (Pearl, 1988). Counting the number of satisfying assignments of a propositional formula (#SAT) is also a well-studied problem that is of fundamental theoretical importance. These two problems are known to be closely related. In particular, the decision versions of both #SAT and BAYES are #P-complete (Valiant, 1979b, 1979a; Roth, 1996), and there are *natural* polynomial-time reductions from each problem to the other (Darwiche, 2002; Sang, Beame, & Kautz, 2005b; Chavira, Darwiche, & Jaeger, 2006).

A more direct relationship between these two problems arises from the observation that they are both instances of the more general "sum of products" problem (SUMPROD). Perhaps the most fundamental algorithm for SUMPROD (developed in a general way by Dechter 1999) is based on the idea of eliminating the variables of the problem one by one following some fixed order. This algorithm is called variable elimination (VE), and it is the core notion in many state-of-the-art exact algorithms for SUMPROD (and BAYES).

SAT, the problem of determining whether or not a propositional formula has any satisfying assignments, is also an instance of SUMPROD, and the original Davis-Putnam algorithm (DP) for determining satisfiability (Davis & Putnam, 1960) which uses ordered resolution is a version of variable elimination. However, DP is never used in practice as its performance is far inferior to modern versions of the backtracking search based DPLL algorithm (Davis, Logemann, & Loveland, 1962). In fact DP is provably less powerful than modern versions of DPLL equipped with clause learning (Hertel, Bacchus, Pitassi, & van Gelder, 2008).

This performance gap naturally raises the question of whether or not backtracking search could be used to solve other types of SUMPROD problems more efficiently than variable elimination. In this paper, we present a general algorithmic framework for using backtrack search methods (specifically DPLL) to solve SUMPROD and related problems.[1] We first show that a straightforward adaptation of backtracking for solving SUMPROD is insufficient. However, by examining the sources of inefficiency we are able to develop some simple caching schemes that allow our backtracking algorithm, #DPLL-Cache, to achieve the same performance guarantees as state-of-the-art exact algorithms for SUMPROD, in terms of both time and space. Furthermore, we prove that backtracking's natural additional flexibility allows it to sometimes achieve an exponential speedup over other existing algorithms. Specifically, we present a family of SUMPROD instances where #DPLL-Cache achieves an exponential speedup over the original versions of three prominent algorithms for SUMPROD.

Besides these theoretical results, there are also good reasons to believe that backtracking based algorithms have the potential to perform much better than their worst case guarantees on problems that arise from real domains. In fact, subsequent work has investigated the practical application of the ideas presented here to the problem of counting satisfying assignments, BAYES, and constraint optimization with very successful results (Sang, Bacchus, Beame, Kautz, & Pitassi, 2004; Sang et al., 2005b; Sang, Beame, & Kautz, 2005a, 2007; Davies & Bacchus, 2007; Kitching & Bacchus, 2008).

An outline of the paper follows. In Section 2, we define SUMPROD; demonstrate that #SAT, BAYES, and other important problems are instances of this class of problems; discuss various graph-theoretic notions of width that can be used to characterize the complexity of algorithms for SUMPROD; and review some core state-of-the-art exact algorithms for SUMPROD. In Section 3, we discuss DPLL-based algorithms with caching for solving #SAT and SUMPROD and provide worst case complexity bounds for these algorithms. These bounds are the same as the best time and space guarantees achieved by currently known algorithms. In Section 4, we provide a framework for comparing our algorithms with other algorithms for SUMPROD and prove that with caching DPLL can efficiently simulate known exact algorithms while sometimes achieving super-polynomially superior performance. In Section 5 we discuss some of the work that has used our algorithmic ideas to build practical solvers for various problems. Finally, we provide some closing remarks in Section 6.

## 2. Background

In this section, we first define the sum-of-products (SUMPROD) class of problems, and then illustrate how BAYES, #SAT, and some other important problems are instances of SUMPROD. As we will show in the rest of the paper, backtracking search equipped with different caching schemes is well suited for solving SUMPROD. The key computational structure that is exploited by all algorithms for SUMPROD is then explained and the graph theoretic notion of width that captures this structure is identified. Different notions of "width" exist, and we present three different definitions and show that they all yield essentially equivalent measures of complexity. The different definitions are however very useful in that different algorithms are most easily analyzed using different definitions of width. Finally, we briefly review some of the most important exact algorithms for solving SUMPROD and related problems.

---

1. The notion of "backtracking" over a previous set of commitments can be utilized in other contexts, including in other algorithms for SUMPROD. However, here we are referring to the standard algorithmic paradigm of backtracking search that explores a single tree of partial variable assignments in a depth-first manner. This algorithm has an extensive history that stretches back over a hundred years (Bitner & Reingold, 1975).

## 2.1 Sum-of-Products

Dechter (1999) has been shown that BAYES and many other problems are instances of a more general problem called SUMPROD (sum-of-products). An instance of SUMPROD is defined by the tuple $\langle \mathcal{V}, \mathcal{F}, \oplus, \otimes \rangle$, where $\mathcal{V}$ is a set of discrete valued variables $\{X_1, \ldots, X_n\}$, $\mathcal{F}$ is a set of functions $\{f_1, \ldots, f_m\}$ with each $f_i$ defined over some set of variables $E_i \subseteq \mathcal{V}$, $\oplus$ is an addition operator, and $\otimes$ is a multiplication operator. The range of the functions in $\mathcal{F}$ depends on the problem, with $\oplus$ and $\otimes$ being operators over that range such that both are commutative, associative, and $\otimes$ distributes over $\oplus$. Typical examples involve functions that range over the boolean domain, with $\oplus$ being disjunction $\vee$ and $\otimes$ being conjunction $\wedge$, or over the reals, with $\oplus$ and $\otimes$ being ordinary addition and multiplication.

**Definition 1** (SUMPROD) Given $\langle \mathcal{V}, \mathcal{F}, \oplus, \otimes \rangle$ the SUMPROD problem is to compute

$$\bigoplus_{X_1} \bigoplus_{X_2} \cdots \bigoplus_{X_n} \bigotimes_{i=1}^{m} f_i(E_i),$$

i.e., the sum ($\oplus$) over all values (assignments) of the variables $\mathcal{V}$ of the product ($\otimes$) of the functions $\mathcal{F}$ evaluated at those assignments.

A number of well known problems are instances of SUMPROD. We describe some of them below.

### 2.1.1 BAYES:

BAYES is the problem of computing probabilities in a Bayesian Network (BN). Developed by Pearl (1988), a Bayesian network is a triple $(\mathcal{V}, E, \mathcal{P})$ where $(\mathcal{V}, E)$ describes a directed acyclic graph, in which the nodes $\mathcal{V} = \{X_1, \ldots, X_n\}$ represent discrete random variables, edges represent direct correlations between the variables, and associated with each random variable $X_i$ is a conditional probability table CPT (or function), $f_i(X_i, \pi(X_i)) \in \mathcal{P}$, that specifies the conditional distribution of $X_i$ given assignments of values to its parents $\pi(X_i)$ in $(\mathcal{V}, E)$. A BN represents a joint distribution over the random variables $\mathcal{V}$ in which the probability of any assignment $(x_1, \ldots, x_n)$ to the variables is given by the equation $Pr(x_1, \ldots, x_n) = \prod_{i=1}^{n} f_i(x_i, \pi(x_i))$, where $f_i(x_i, \pi(x_i))$ is $f_i$ evaluated at this particular assignment.

The generic BAYES problem is to compute the posterior distribution of a variable $X_i$ given a particular assignment to some of the other variables $\alpha$: i.e., $Pr(X_i|\alpha)$. Since $X_i$ has only a finite set of $k$ values, this problem can be further reduced to that of computing the $k$ values $Pr(X_i = d_j \wedge \alpha)$, $j = 1, \ldots, k$ and then normalizing them so that they sum to 1. The values $Pr(X_i = d_j \wedge \alpha)$ can be computed by making all of the assignments in $\alpha$ as well as $X_i = d_j$, and then summing out the other variables from the joint distribution $Pr(x_1, \ldots, x_n)$. Given the above product decomposition of $Pr(x_1, \ldots, x_n)$, this is equivalent to reducing the functions $f_i \in \mathcal{P}$ by setting the variables assigned in $\alpha$ and $X_i = d_j$, and then summing their product over the remaining variables; i.e., it is an instance of SUMPROD.

**Computing all Marginals** It is common when solving BAYES to want to compute all marginals. That is, instead of wanting to compute just the marginal $Pr(X_i|\alpha)$ for one particular variable $X_i$, we want to compute the marginal for all variables not instantiated by $\alpha$.

### 2.1.2 MARKOV RANDOM FIELDS

Markov Random Fields or Markov Networks (MN) (Preston, 1974; Spitzer, 1971) are similar to Bayesian Networks in that they also define a joint probability distribution over a set of discrete random variables $\mathcal{V} = \{X_1, \ldots, X_n\}$ using a set of functions $f_i$, called potentials, each over some set of variables $E_i \subseteq \mathcal{V}$. In particular, the probability of any assignment $(x_1, \ldots, x_n)$ to the variables is given by the normalized product of the $f_i$ evaluated at the values specified by the assignment: $\prod_i f_i(E_i[x_1, \ldots, x_n])$. The difficulty is to compute the partition function, or normalizing constant:

$$Z = \sum_{X_1} \cdots \sum_{X_n} \prod_{i=1}^{m} f_i(E_i).$$

Computing the partition function is thus an instance of SUMPROD.

### 2.1.3 MOST PROBABLE EXPLANATION

Most Probable Explanation (MPE) is the problem of finding the most probable complete assignment to the variables in a Bayes net (or Markov net) that agrees with a fixed assignment to a subset of the variables (the evidence). If the evidence, $\alpha$, is an instantiation of the variables in $E \subset \mathcal{V}$, then MPE is the problem of computing

$$\max_{V-E} \prod_{i=1}^{m} f_i|_\alpha (E_i - E),$$

where $f_i|_\alpha$ is the reduction of the function $f_i$ by the instantiations $\alpha$ to the variables in $E$ (yielding a function over the variables $E_i - E$).

### 2.1.4 SAT

Let $\mathcal{V} = \{X_1, X_2, \ldots, X_n\}$ be a collection of $n$ Boolean variables, and let $\phi(\mathcal{V})$ be a $k$-CNF Boolean formula on these variables with $m$ clauses $\{c_1, \ldots, c_m\}$. An assignment $\alpha$ to the Boolean variables $\mathcal{V}$ is *satisfying* if it makes the formula True (i.e., $\phi(\alpha) = 1$). SAT asks, given a Boolean formula $\phi(\mathcal{V})$ in $k$-CNF, does it have a satisfying assignment? By viewing each clause $c_i$ as being a function of its variables $E_i$ (i.e., it maps an assignment to these variables to TRUE if that assignment satisfies the clause and to FALSE otherwise), we can see that SAT is equivalent to the instance of SUMPROD $\langle \mathcal{V}, \{c_1, \ldots, c_m\}, \vee, \wedge \rangle$:

$$\bigvee_{X_1} \cdots \bigvee_{X_n} \bigwedge_{i=1}^{m} c_i(E_i).$$

### 2.1.5 #SAT

Given a $k$-CNF formula $\phi(\mathcal{V})$ on the boolean variables $\mathcal{V} = \{X_1, \ldots, X_n\}$, as above, #SAT is the problem of determining the number of satisfying assignments for $\phi$. By viewing each clause $c_i$ as being a function from its variables $E_i$ to $\{0, 1\}$ (i.e., it maps satisfying assignments to 1 and falsifying assignments to 0), we can see that #SAT is equivalent to the instance of SUMPROD $\langle \mathcal{V}, \{c_1, \ldots, c_m\}, +, \times \rangle$:

$$\sum_{X_1} \cdots \sum_{X_n} \prod_{i=1}^{m} c_i(E_i).$$

### 2.1.6 OPTIMIZATION WITH DECOMPOSED OBJECTIVE FUNCTIONS

Let $\mathcal{V} = \{X_1, \ldots, X_n\}$ be a collection of finite valued variables, the optimization task is to find an assignment of values to these variables that maximizes some objective function $O(\mathcal{V})$ (i.e., a function that maps every complete assignment to the variables to a real value). In many problems $O$ can be decomposed into a sum of sub-objective functions $\{f_1, \ldots, f_m\}$ with each $f_i$ being a function of some subset of the variables $E_i$. This problem can then be cast as the SUMPROD instance $\langle \mathcal{V}, \{f_1, \ldots, f_m\}, \max, + \rangle$

$$\max_{X_1} \cdots \max_{X_n} \sum_{i=1}^{m} f_i(E_i).$$

## 2.2 The Computational Complexity of SUMPROD

SUMPROD is a computationally difficult problem. For example, #SAT is known to be complete for the complexity class #P (Valiant, 1979b, 1979a) as is BAYES (Roth, 1996). Many special cases that are easy for SAT remain hard for #SAT, e.g., Valiant showed that the decision version of #SAT is #P hard even when the clause size, $k$, is 2, and Roth (1996) showed that the problem is hard to even approximate in many cases where SAT is easy, e.g., when $\phi(\mathcal{V})$ is monotone, or Horn, or 2-CNF.

Despite this worst case intractability, algorithms for SUMPROD, e.g., the variable elimination algorithm presented by Dechter (1999), can be successful in practice. The key structure exploited by this algorithm, and by most algorithms, is that the functions $f_i$ of many SUMPROD problems are often relatively local and fairly independent. That is, it is often the case that the sets of variables $E_i$ that each function $f_i$ depends on are small, so that each function is dependent only on a small "local" set of the variables, and that these sets share only a few variables with each other, so that the functions $f_i$ are fairly independent of each other. The graph theoretic notion of Tree Width is used to make these intuitions precise.

## 2.3 Complexity Measures and Tree width

There is a natural hypergraph, $\mathcal{H} = (V, E)$, corresponding to any instance $\langle \mathcal{V}, \mathcal{F}, \oplus, \otimes \rangle$ of SUMPROD. In the hypergraph, $V$ corresponds to the set $\mathcal{V}$ of variables, and for every function $f_i$ with domain set $E_i$, there is a corresponding hyperedge, $E_i$.

The "width" of this hypergraph is the critical measure of complexity for essentially all state-of-the-art algorithms for #SAT, BAYES, and SUMPROD. There are three different (and well known) notions of width that we will define in this section. We will also show that these different notions of width are basically equivalent. These equivalences are known, although we need to state them and prove some basic properties, in order to analyze our new algorithms, and to relate them to standard algorithms.

**Definition 2** (**Branch width**) Let $\mathcal{H} = (V, E)$ be a hypergraph. A **branch decomposition** of $\mathcal{H}$ is a binary tree $T$ such that each node of $T$ is labelled with a subset of $V$. There are $|E|$ many leaves of $T$, and their labels are in one-to-one correspondence with the hyperedges $E$. For any other node $n$ in $T$, let $A$ denote the union of the leaf labeling of the subtree rooted at $n$, and let $B$ denote the union of the labelings of the rest of the leaves. Then the label for $n$ is the set of all vertices $v$ that are in the intersection of $A$ and $B$. The **branch width** of a branch decomposition $T$ for $\mathcal{H}$ is the maximum size of any labeling in $T$. The **branch width** of $\mathcal{H}$ is the minimum branch width over all branch decompositions of $\mathcal{H}$.

**Example 1** Figure 1 shows a particular branch decomposition $T_{bd}$ for the hypergraph $\mathcal{H} = (V, E)$ where $V = \{1, 2, 3, 4, 5\}$ and $E = \{\{1, 2, 3\}, \{1, 4\}, \{2, 5\}, \{3, 5\}, \{4, 5\}\}$. $T_{bd}$ has branch width 3.

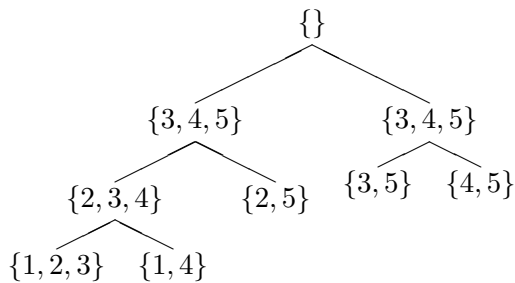

Figure 1: A branch decomposition of branch width 3 for $\mathcal{H} = \{(1, 2, 3), (1, 4), (2, 5), (3, 5), (4, 5)\}$.

**Definition 3** (**Elimination width**) Let $\mathcal{H} = (V, E)$ be a hypergraph, and let $\pi = v_1^\pi, \ldots, v_n^\pi$ be an ordering of the vertices in $V$, where $v_i^\pi$ is the $i^{th}$ element in the ordering. This induces a sequence of hypergraphs $\mathcal{H}_n, \mathcal{H}_{n-1}, \ldots, \mathcal{H}_1$ where $\mathcal{H} = \mathcal{H}_n$ and $\mathcal{H}_{i-1}$ is obtained from $\mathcal{H}_i$ as follows. All edges in $\mathcal{H}_i$ containing $v_i^\pi$ are merged into one edge and then $v_i^\pi$ is removed. Thus the underlying vertices of $\mathcal{H}_i$ are $v_1^\pi, \ldots v_{i-1}^\pi$. The **induced width** of $\mathcal{H}$ under $\pi$ is the size of the largest edge in all the hypergraphs $\mathcal{H}_n, \ldots, \mathcal{H}_1$. The **elimination width** of $\mathcal{H}$ is the minimum induced width over all orderings $\pi$.

**Example 2** Under the ordering $\pi = \langle 1, 2, 3, 4, 5 \rangle$ the hypergraph $\mathcal{H}$ of Example 1 produces the following sequence of hypergraphs:

$$
\begin{aligned}
\mathcal{H}_5 &= \{(1, 2, 3), (1, 4), (2, 5), (3, 5), (4, 5)\} \\
\mathcal{H}_4 &= \{(2, 3, 4), (2, 5), (3, 5), (4, 5)\} \\
\mathcal{H}_3 &= \{(3, 4, 5), (3, 5), (4, 5)\} \\
\mathcal{H}_2 &= \{(4, 5), (4, 5)\} \\
\mathcal{H}_1 &= \{(5)\}
\end{aligned}
$$

The induced width of $\mathcal{H}$ under $\pi$ is 3—the edges $(1, 2, 3) \in \mathcal{H}_1$, $(2, 3, 4) \in \mathcal{H}_2$ and $(3, 4, 5) \in \mathcal{H}_3$ all achieve this size.

Tree width is the third notion of width.

**Definition 4 (Tree width)** Let $\mathcal{H} = (V, E)$ be a hypergraph. A **tree decomposition** of $\mathcal{H}$ is a binary tree $T$ such that each node of $T$ is labelled with a subset of $V$ in the following way. First, for every hyperedge $e \in E$, some leaf node in $T$ must have a label that contains $e$. Secondly, given labels for the leaf nodes every internal node $n$ contains $v \in V$ in its label if and only if $n$ is on a path between two leaf nodes $l_1$ and $l_2$ whose labels contain $v$.[2] The **tree width** of a tree decomposition $T$ for $\mathcal{H}$ is the maximum size of any labeling in $T$ minus 1, and the **tree width** of $\mathcal{H}$ is the minimum tree width over all tree decompositions of $\mathcal{H}$.

**Example 3** Figure 2 shows $T_{td}$ a tree decomposition for $\mathcal{H}$ of Example 1. $T_{td}$ has tree width 3.

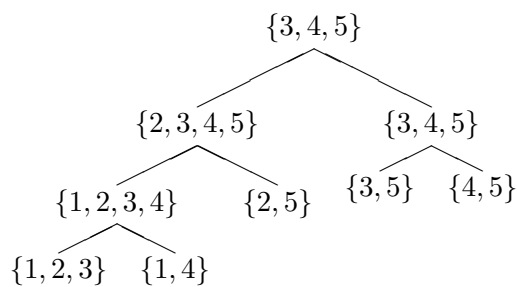

Figure 2: Tree decomposition of tree width for 3 for $\mathcal{H}$ of Example 1.

The next three lemmas show that these three notions are basically equivalent. The proofs of Lemmas 2 and 3 are given in the appendix.

**Lemma 1 (Robertson & Seymour, 1991)** Let $\mathcal{H}$ be a hypergraph. Then the branch width of $\mathcal{H}$ is at most the tree width of $\mathcal{H}$ plus 1, and the tree width of $\mathcal{H}$ is at most 2 times the branch width of $\mathcal{H}$.

**Lemma 2** Let $\mathcal{H} = (V, E)$ be a hypergraph with a tree decomposition of width $w$. Then there is an elimination ordering $\pi$ of the vertices $V$ such that the induced width of $\mathcal{H}$ under $\pi$ is at most $w$.

**Lemma 3** Let $\mathcal{H}$ be a hypergraph with elimination width at most $w$. Then $\mathcal{H}$ has a tree decomposition of tree width at most $w$.

Letting $TW(\mathcal{H})$, $BW(\mathcal{H})$, and $EW(\mathcal{H})$ represent the tree width, branch width and elimination width of the hypergraph $\mathcal{H}$, the above lemmas give the following relationship between these three notions of width: for all hypergraphs $\mathcal{H}$

$$BW(\mathcal{H}) - 1 \leq TW(\mathcal{H}) = EW(\mathcal{H}) \leq 2BW(\mathcal{H}).$$

---

2. Since the labels of internal nodes are determined by the labels of the leaf nodes in this way, it can be seen that for any pair of nodes $n_1$ and $n_2$ in the tree decomposition every node lying on the path between them must contain $v$ in its label if $v$ appears in both $n_1$'s and $n_2$'s labels. This is commonly known as the running intersection property of tree decompositions.

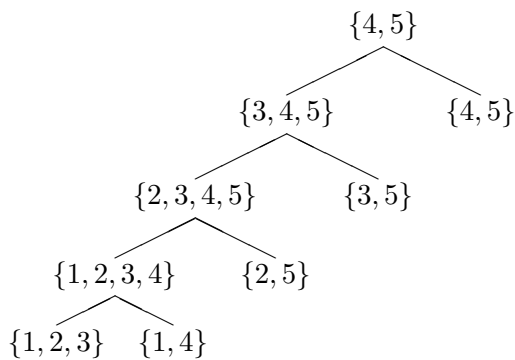

Figure 3: Tree decomposition of the hypergraph $\mathcal{H}$ of Example 1 that has been constructed from the ordering $\pi = \langle 1, 2, 3, 4, 5 \rangle$.

**Example 4** The tree decomposition $T_{td}$ of $\mathcal{H} = \{(1, 2, 3), (1, 4), (2, 5), (3, 5), (4, 5)\}$ given in Figure 2 has the property that it has tree width no more than twice the branch width of the branch decomposition $T_{bd}$ of $\mathcal{H}$ given in Figure 1. From $T_{td}$ we can obtain the ordering $\pi = \langle 1, 2, 3, 4, 5 \rangle$ that was used in Example 2. (The proof of Lemma 2, given in the appendix, shows how a elimination ordering can be constructed from a tree-decomposition.) As shown in Example 2, $\pi$ has induced width 3, equal to the tree width of tree decomposition $T_{td}$ from which it was constructed. Finally, from the ordering $\pi$ we can construct a new tree decomposition for $\mathcal{H}$ shown in Figure 3. (The proof of Lemma 3 shows how a tree decomposition can be constructed from an elimination ordering). $\pi$ has induced width 3 and, as indicated by Lemma 3 the tree decomposition constructed from it has equal tree width of 3.

It can be noted that our definition of tree decompositions varies slightly from other definitions that appear in the literature, e.g., (Bodlaender, 1993). Following Robertson and Seymour (1991) we have defined tree decompositions over hypergraphs, rather than over graphs, and we have made two extra restrictions so as to simplify the proofs of our results. First, we have restricted tree decompositions to be binary trees, and second we have required that each hyperedge be contained in the label of some *leaf* node of the tree decomposition. Usually tree decompositions are not restricted to be binary trees, and only require that each hyperedge be contained in some node's label (not necessarily a leaf node).

It is not difficult to show that any tree decomposition that fails to satisfy our two restrictions can be converted to a tree decomposition satisfying these restrictions without changing its width. However, it is more straight forward to observe that with or without these two restrictions tree width is equal to elimination width. Hence, our restrictions do not change the tree width.

### 2.4 Exact Algorithms for SUMPROD

Next we briefly review three prominent exact algorithms for BAYES. These algorithms solve the more general problem SUMPROD. All of these algorithms are in fact nondeterministic algorithms that should be considered to be families of procedures, each member of which is a particular deterministic realization.

### 2.4.1 VARIABLE ELIMINATION:

Variable or bucket elimination (VE) (Dechter, 1999) is a fundamental algorithm for SUMPROD. Variable elimination begins by choosing an elimination ordering, $\pi$ for the variables $\mathcal{V} = \{X_1, \ldots, X_n\}$: $X_{\pi(1)}, \ldots, X_{\pi(n)}$. (This is the nondeterministic part of the computation). In the first phase, all functions involving $X_{\pi(1)}$, are collected together in the set $\mathcal{F}_{X_{\pi(1)}}$, and a new function, $F_1$ is computed by "summing out" $X_{\pi(1)}$. The new function sums the product of all the functions in $\mathcal{F}_{X_{\pi(1)}}$ over all of $X_{\pi(1)}$'s values. Specifically, $F_1$ is a function of all of the variables of the functions in $\mathcal{F}_{X_{\pi(1)}}$ except for $X_{\pi(1)}$, and its value on any assignment $\alpha$ to these variables is

$$F_1(\alpha) = \sum_{d \in \text{vals}(X_{\pi(1)})} \prod_{f \in \mathcal{F}_{X_{\pi(1)}}} f(\alpha, X_{\pi(1)} = d).$$

Summing out $X_{\pi(1)}$ induces a new hypergraph, $\mathcal{H}_1$, where the hyperedges corresponding to the set of functions $\mathcal{F}_{X_{\pi(1)}}$ are replaced by a single hyperedge corresponding to the new function $F_1$. The process then continues to sum out $X_{\pi(2)}$ from $\mathcal{H}_1$ and so on until all $n$ variables are summed out. Note that the sequence of hypergraphs generated by summing out the variables according to $\pi$ is the same the sequence of hypergraphs that defines the induced width of $\pi$ (Definition 3).

The original Davis-Putnam algorithm (Davis & Putnam, 1960) based on ordered resolution is an instance of variable elimination. Consider applying variable elimination to the formulation of SAT given above. For SAT, the new functions $F_i$ computed at each stage need only preserve whether or not the product of the functions in $\mathcal{F}_{X_{\pi(i)}}$ is 0 or 1, the exact number of satisfying assignments need not remembered. This can be accomplished by representing the $F_i$ symbolically as a set of clauses. Furthermore, this set of clauses can be computed by generating all clauses that can be obtained by resolving on $X_{\pi(i)}$, and then discarding all old clauses containing $X_{\pi(i)}$. This resolution step corresponds to the summing out operation, and yields precisely the Davis-Putnam (DP) algorithm for satisfiability.[3]

### 2.4.2 RECURSIVE CONDITIONING:

Recursive conditioning (RC) (Darwiche, 2001) is another type of algorithm for SUMPROD. Let $\mathcal{S} = \langle \mathcal{V}, \mathcal{F}, \oplus, \otimes \rangle$ be an instance of SUMPROD and $\mathcal{H}$ be its underlying hypergraph. RC is a divide and conquer algorithm that instantiates the variables of $\mathcal{V}$ so as to break the problem into disjoint components. It then proceeds to solve these components independently. The original space-efficient version of recursive conditioning, as specified by Darwiche (2001), begins with a branch decomposition $T$ of $\mathcal{H}$ of width $w$ and depth $d$, and an initially empty set of instantiated variables $\rho$. (Choosing $T$ is the nondeterministic part of the computation.) We call this algorithm **RC-Space** and show it in Algorithm 1.

The branch decomposition $T$ specifies a recursive decomposition of the problem and is used by RC-Space as follows. Let $label(n)$ be the label of a node in $T$, and let $\mathcal{S}_T$ be the SUMPROD problem defined by the variables and functions contained in $T$. (In the initial call $T$ is the complete branch decomposition containing all variables and functions of $\mathcal{S}$, so that initially $\mathcal{S}_T = \mathcal{S}$). Starting at $r$, the root of $T$, RC-Space solves the reduced SUMPROD $\mathcal{S}_T|_{\rho \cup \alpha}$ for all assignments $\alpha$ to the variables in $label(left(r)) \cap label(right(r))$ not yet instantiated by $\rho$, where $left(r)$ and $right(r)$ are the left and right children of $r$. The sum over all such $\alpha$ is the solution to the inputed instance $\mathcal{S}_T|_\rho$.

---

3. Rish and Dechter (2000) have previously made a connection between DP and variable elimination. They were thus able to show, that DP runs in time $n^{O(1)} 2^{O(w)}$, where $w$ is the branch width of the underlying hypergraph of the SAT instance.

Each $\alpha$ renders the set of functions in the subtree below $leftChild(r)$ (i.e., the leaf labels) disjoint from the functions below $rightChild(r)$. Thus for each $\alpha$, RC-Space can independently solve the subproblems specified by $leftChild(r)|_{\rho \cup \alpha}$ and $rightChild(r)|_{\rho \cup \alpha}$ (i.e., the sum of the products of all of the functions below the left/right subtree conditioned on the instantiations in $\rho \cup \alpha$) and multiply their answers to obtain the solution to $\mathcal{S}_T|_{\rho \cup \alpha}$. At the leaf nodes, the function $f_i$ associated with that node has had all of its variables instantiated, so the algorithm can simply "LOOKUP" $f_i$'s current value.

---

**Algorithm 1: RC-Space**—Linear Space Recursive Conditioning

---

1 **RC-Space** $(T, \rho)$
2 **begin**
3     **if** $T$ *is a leaf node* **then**
4         **return** *LOOKUP(value of function labeling the leaf node)*
5     $p = 0; r = root(T)$
6     $\vec{x} =$ variables in $label(left(r)) \cap label(right(r))$ uninstantiated by $\rho$
7     **forall** $\alpha \in \{$*instantiations of* $\vec{x}\}$ **do**
8         $p = p +$ **RC-Space** $(leftChild(T), \rho \cup \alpha) \times$ **RC-Space** $(rightChild(T), \rho \cup \alpha)$
9     **end**
10     **return** $p$
11 **end**

---

A less space-efficient but more time-efficient version of recursive conditioning, called **RC-Cache**, caches intermediate values that can be reused to reduce the computation. Algorithm 2 shows the RC-Cache algorithm. Like RC-Space, each invocation of RC-Cache solves the subproblem specified by the variables and functions contained in the passed subtree $T$. Since the functions below $T$ only share the variables in $label(root(T))$ with variables outside of $T$, only the instantiations in the subset, $y$, of $\rho$ intersecting $label(root(T))$ can affect the form of this subproblem. Hence, RC-Cache will return the same answer if invoked with the same $T$ and same $y$, even if other assignments in $\rho$ have changed. RC-Cache, can thus use $T$ and $y$ to index a cache, storing the computed result in the cache (line 13) and returning immediately if the answer is already in the cache (line 7).

**Propagation** Since RC instantiates the problem's variables, propagation can be employed. That is, RC can perform additional inference to compute some of the implicit effects each assignment has on the remaining problem $\mathcal{S}_T|_{\rho \cup \alpha}$. For example, if the functions of the SUMPROD problem are all clauses (e.g., when solving #SAT) unit propagation can be performed. Propagation can make recursive conditioning more effective. For example, if one of the remaining clauses becomes falsified through unit propagation, recursive conditioning can immediately move on to the next instantiation of the variables $\vec{x}$. Similarly, unit propagation can force the value of variables that will be encountered in subsequent recursive calls, thus reducing the number of different instantiations $\alpha$ that must be attempted in that recursive call. It can be noted that propagation does not reduce the worst case complexity of the algorithm, as on some SUMPROD problems propagation is ineffective. It can however improve the algorithm's efficiency on some families of problems.

---

**Algorithm 2**: **RC-Cache**—Recursive Conditioning with caching

**1** **RC-Cache** $(T, \rho)$
**2** **begin**
**3**    **if** $T$ *is a leaf node* **then**
**4**       **return** *LOOKUP(value of function labeling the leaf node)*
**5**    $\mathbf{y} = \rho \cap label(root(T))$
**6**    **if** *InCache*$(T, \mathbf{y})$ **then**
**7**       **return** *GetValue*$(T, \mathbf{y})$
**8**    $p = 0; r = root(T)$
**9**    $\vec{x} =$ variables in $label(left(r)) \cap label(right(r))$ uninstantiated by $\rho$
**10**    **forall** $\alpha \in \{$*instantiations of* $\vec{x}\}$ **do**
**11**       $p = p +$ **RC-Cache** $(leftChild(T), \rho \cup \alpha) \times$ **RC-Cache** $(rightChild(T), \rho \cup \alpha)$
**12**    **end**
**13**    AddToCache$((T, \mathbf{y}), p)$
**14**    **return** $p$
**15** **end**

---

**RC-Cache$^+$**   A simple extension of RC that is used in practice is to set the variables $\vec{x} \subseteq label(left(r)) \cap label(right(r))$ (line 10 of Algorithm 2) iteratively rather than all at once. That is, rather than iterate over all complete assignments $\alpha$ to $\vec{x}$ we can instantiate these variables one at a time, performing propagation after each assignment. This can make propagation more effective, since, e.g., an empty clause might be detected after instantiating only a subset of the variables in $\vec{x}$ and thus the number of iterations of the for loop might be reduced.

Once the variables of $\vec{x}$ are being set iteratively the order in which they are assigned can vary. Furthermore, the order of assignment can vary dynamically. That is, depending on how the values assigned to the first $k$ variables of $\vec{x}$, the algorithm can make different choices as to which unassigned variable of $\vec{x}$ to assign next.

We call the extension of RC-Cache that uses incremental assignments and dynamic variable ordering within set $\vec{x}$, **RC-Cache$^+$** . That is RC-Cache$^+$ uses the same caching scheme as RC-Cache, but has more flexibility in its variable ordering. It should be noted however, that RC-Cache$^+$ does not have complete freedom in its variable ordering. It must still follow the inputed branch decomposition $T$. That is, the variable chosen must come from the set $\vec{x} \subseteq label(left(r)) \cap label(right(r))$. This is in contrast with the DPLL based algorithms we present in the next section, which are always free to choose any remaining unassigned variable as the next variable to assign.

**Space-Time Tradeoff**   RC has the attractive feature that it can achieve a non-trivial space-time tradeoff, taking less time if it caches its recursively computed values (RC-Cache), or taking less space without caching (RC-Space). In fact, Darwiche and Allen (2002) show that there is a smooth tradeoff that can be achieved, with RC-Space and RC-Cache at the two extremes.

The DPLL based algorithms presented here share a number of features with RC; they also reduce and decompose the input problem by making instantiations, gain efficiency by caching, and achieve a similar space-time tradeoff. However, our algorithms are based on the paradigm of backtracking, rather than divide and conquer. In particular, they explore a single backtracking tree in which the decomposed subproblems are not solved separately but rather can be solved in any interleaved fashion. As a result, they are not limited to following the decomposition scheme specified by a fixed branch decomposition. As we will see, the limitation of a static decomposition scheme means that RC-Space and RC-Cache must perform exponentially worse than our algorithms on some instances.

### 2.4.3 AND/OR SEARCH:

In more recent work Dechter and Mateescu (2007) have shown that the notion of AND/OR search spaces (Nilsson, 1980) can be applied to formalize the divide and conquer approach to SUMPROD problems utilized by RC. In this formulation the structure that guides the AND/OR search algorithm is a pseudo tree. (Choosing the pseudo tree is the nondeterministic part of the computation.)

**Definition 5** (**Primal Graph**) The **primal graph** of a hypergraph $\mathcal{H}$ is an undirected graph $G$ that has the same vertices as $\mathcal{H}$ and has an edge connecting two vertices if and only if those two vertices appear together in some hyperedge of $\mathcal{H}$.

**Definition 6** (**Pseudo Tree**) Given an undirected graph $G$ with vertices and edges $(V, E_G)$, a **pseudo tree** for $G$ is a directed rooted tree $T$ with vertices and edges $(V, E_T)$ (i.e., the same set of vertices as G), such that any edge $e$ that is in $G$ but not in $T$ must connect a vertex in $T$ to one of its ancestors. That is, $e = (v_1, v_2) \wedge e \in E_G \wedge e \notin E_T$ implies that either $v_1$ is an ancestor of $v_2$ in $T$ or $v_2$ is an ancestor of $v_1$ in $T$.

This implies that there is no edge of $G$ connecting vertices lying in different subtrees of $T$. Given a SUMPROD problem $\mathcal{S} = \langle \mathcal{V}, \mathcal{F}, \oplus, \otimes \rangle$ with underlying hypergraph $\mathcal{H}$, we can form $G$, the primal graph of $\mathcal{H}$. The vertices of $G$ are the variables of the problem $\mathcal{V}$ and any pair of variables that appear together in some function of $\mathcal{F}$ will be connected by an edge in $G$. A pseudo tree $T$ for $G$ will then have the property that two vertices of $T$ (variables of $\mathcal{S}$) can only appear in functions of $\mathcal{F}$ with their ancestors or their descendants, they cannot appear in functions with their siblings nor with their ancestor's siblings nor with the descendants of such siblings.

This implies that once a variable $v$ and all of its ancestors in $T$ have been instantiated, the variables contained in its children subtrees become disconnected. That is, the variables in these subtrees no longer appear in functions together, and the resulting subproblems can be solved independently. The AND/OR search algorithm utilizes this fact to solve these subproblems independently, just like recursive conditioning.

**Example 5** Given the hypergraph $\mathcal{H} = (V, E)$ where $V = \{1, 2, 3, 4, 5\}$ and $E = \{\{1, 2, 3\}, \{1, 4\}, \{2, 5\}, \{3, 5\}\}$, the primal graph of $\mathcal{H}$ is $G = (V, E_G)$ where $E_G = \{(1, 2), (1, 3), (2, 3), (1, 4), (2, 5), (3, 5)\}$. $\mathcal{H}$, its primal graph $G$, and a pseudo tree for $G$ are shown in Figure 4. The dotted lines shown on the pseudo tree are the edges of $G$ that are not in the pseudo tree. As can be seen from the diagram these edges connect nodes only with their ancestors.

The space efficient version of the **AND/OR-Space** search algorithm (Dechter & Mateescu, 2007) is shown in Algorithm 3. It solves the SUMPROD instance $\mathcal{S} = \langle \mathcal{V}, \mathcal{F}, \oplus, \otimes \rangle$, taking as input a pseudo tree for the problem $T$ (i.e., the hypergraph for $\mathcal{S}$ is converted to a primal graph $G$, and $T$ is a pseudo tree for $G$), and an initially empty set of instantiated variables $\rho$. The algorithm solves a sub-problem of the original instance $\mathcal{S}$ reduced by the instantiations $\rho$, $\mathcal{S}|_\rho$. The sub-problem being solved is defined by the functions of $\mathcal{S}|_\rho$ that are over the variables contained in the passed sub-tree $T$. Initially, with $\rho$ being empty and $T$ being the original pseudo tree containing all variables, the algorithm solves the original problem $\mathcal{S}$.

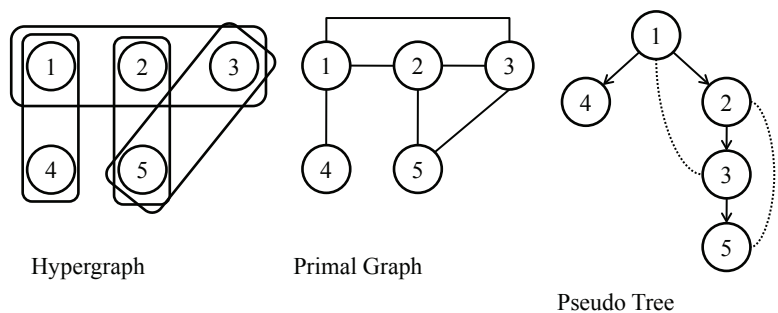

Figure 4: The hypergraph, primal graph, and a pseudo tree for Example 5.

The nodes of the pseudo tree $T$ are variables of the problem $\mathcal{S}$, and we also attach to each node $n$ of $T$ a set of functions $fns(n)$. A function $f$ of $\mathcal{F}$ is in $fns(n)$ if and only if (a) $n$ is in the scope of $f$ and (b) all other variables in the scope of $f$ are ancestors of $n$ in $T$. This means that $f$ will have a fully instantiated set of arguments when AND/OR search instantiates the node (variable) $n$.

---

**Algorithm 3**: **AND/OR-Space**—Linear Space AND/OR search

---

**1** **AND/OR-Space** $(T, \rho)$
**2** **begin**
**3**     $p = 0; r = root(T)$
**4**     $ST_r =$ set of subtrees below $r$
**5**     **forall** $d \in \{instantiations\ of\ r\}$ **do**
**6**         $\alpha = \prod_{f \in fns(r)}$ LOOKUP(value of $f$ on $\rho \cup \{r = d\}$)
**7**         $p = p + \alpha \times \prod_{T' \in ST_r}$ **AND/OR-Space** $(T', \rho \cup \{r = d\})$
**8**     **end**
**9**     **return** $p$
**10** **end**

---

The algorithm operates on the variable $r$ that is the root of the pseudo tree $T$. For each instantiation of $r$ the algorithm computes $\alpha$, the product of the functions in $\mathcal{F}$ that have now become fully instantiated by the assignment to $r$, i.e., those in $fns(r)$. It then invokes a separate recursion for each child of $r$ passing the subtree rooted by that child to the recursive call. AND/OR search exploits decomposition through these separate recursions. If $r$ has only one child, then the problem is not decomposed—there is only the single reduced subproblem that has resulted from instantiating $r$.

Like RC, AND/OR search can be made more time efficient at the expense of using more space. Algorithm 4 shows the caching version **AND/OR-Cache** (called AND/OR graph search by Dechter and Mateescu (2007)). Let $label(n)$ for any node $n$ in the pseudo tree $T$ be the set of ancestors of $n$ that appear in some function with $n$ or with some descendant of $n$ in $T$. It is only the instantiations to $label(n)$ that can affect the functions over the variables in the subtree rooted by $n$. Hence, $label(n)$ plays the same role as the root label of the passed branch decomposition in RC-Cache: only instantiations to these variables can affect the subproblem currently being computed.

Hence, like RC-Cache, AND/OR-Cache can use the instantiations in the subset, $y$, of $\rho$ intersecting $label(root(T))$ along with $T$ to index a cache.

Finally, as with RC-Cache$^+$, propagation can be used to decrease the number of branches that AND/OR search needs to explore. For example, the recursive calls over the children of $r$ can be terminated when one of these calls returns the value zero.

---

**Algorithm 4**: **AND/OR-Cache**—AND/OR search with caching

---

**1** **AND/OR-Cache** $(T, \rho)$
**2** **begin**
**3**   $p = 0; r = root(T)$
**4**   $\mathbf{y} = \rho \cap label(root(T))$
**5**   **if** $InCache(T, \mathbf{y})$ **then**
**6**     **return** $GetValue(T, \mathbf{y})$
**7**   $ST_r = $ set of subtrees below $r$
**8**   **forall** $d \in \{instantiations\ of\ r\}$ **do**
**9**     $\alpha = \prod_{f \in fns(r)}$ LOOKUP(value of $f$ on $\rho \cup \{r = d\}$)
**10**     $p = p + \alpha \times \prod_{T' \in ST_r}$ **AND/OR-Cache** $(T', \rho \cup \{r = d\})$
**11**   **end**
**12**   **return** $p$
**13** **end**

---

**AND/OR-Cache$^+$**    Some variable order dynamism can be employed during AND/OR search. In particular, the variables along any chain in the pseudo tree $T$ can be reordered without affecting the decompositions specified by $T$. A chain is a sub-path of $T$ such that none of its nodes, except perhaps the last, have more than one child. In Figure 4 nodes 2, 3, and 5 form a chain. The resultant extension, **AND/OR-Cache$^+$**, can dynamically chose to next instantiate any of the variables in the chain that starts at the root of its passed pseudo tree $T$. (Marinescu and Dechter (2006) refer to AND/OR-Cache$^+$ as "AND/OR with partial variable ordering". However they did not utilize caching in their version of the algorithm.)

It will then pass the rest of the chain (and the nodes below) to its next recursive call, or if the chosen variable was the last in the chain it will invoke a separate recursive call for each child. Like RC-Cache$^+$, AND/OR-Cache$^+$ does not have complete freedom in its choice of variable—it must chose a variable from the top most chain. Furthermore, AND/OR-Cache$^+$ can only use its caching scheme at the bottom of each chain (i.e., after all variables in the chain have been instantiated) since its cache requires that the same set of variables be instantiated. This makes AND/OR-Cache$^+$ very similar to RC-Cache$^+$.

### 2.4.4 OTHER EXACT ALGORITHMS

The algorithm most commonly used for BAYES is the **join tree** algorithm (Lauritzen & Spiegelhalter, 1988), which can also be adapted to solve other kinds of SUMPROD problems. The join-tree algorithm first organizes the primal graph of the SUMPROD problem into a tree by clustering the variables, and it then performs message passing on the tree where the messages are computed by a variable elimination process. In the context of BAYES the main advantage of join-tree algorithms is that they compute all marginals. That is they compute the posterior probability of all of the variables given some evidence.

In contrast, the default version of variable elimination computes only the posterior distribution for a single variable. However, Kask et al. (2005) show how the join-tree algorithm can be reduced to a version of VE that remembers some of its intermediate results and runs in the same time and space as VE. Hence, all of the results we state here comparing VE with our new backtracking based algorithms also hold for the join tree algorithm.

**Computing all Marginals** All of the algorithms described above, i.e., VE, RC, and AND/OR search, can be modified to compute all marginals when solving BAYES without any change to their worst case complexity. In particular, besides the results of Kask et al. (2005), Darwiche (2001) has shown that RC can compute all marginals on BAYES problems with an extra bottom up traversal of its search tree—at most doubling its run time. The same technique can be applied to AND/OR search algorithms. For the DPLL algorithms we present here, Sang et al. (2005b) have given an even simpler scheme for modifying them so that they can computing all marginals. Sang et al.'s scheme involves maintaining some extra information during search and does not require an extra traversal of the search tree.

Another algorithm that has now been mostly superseded is **cut-set conditioning** (Pearl, 1988). Here the idea is to identify a subset of variables which when set reduce the underlying hypergraph of the SUMPROD into a tree. The reduced SUMPROD can then be easily solved. However, the approach requires trying all possible instantiations of the cut-set yielding a runtime that is usually worse than RC-Cache. Nevertheless, cutset conditioning can potentially be applied in conjunction with other exact algorithms (Mateescu & Dechter, 2005).

Finally, an important early algorithm called **DDP** was presented by Bayardo and Pehoushek (2000). This was a version of DPLL that utilized dynamic decomposition for solving #SAT. In terms of the algorithms discussed above, AND/OR-Space can be viewed as being an version of DDP that utilizes a pseudo tree to guide its variable ordering. In the original presentation of DDP, any variable ordering could be used including dynamic variable orderings. The search continued until the problem was decomposed into independent components (tested for during search) at which point a separate recursion was used to solve each component. Hence, the DDP explored an AND/OR search tree, however this tree need not correspond to any pseudo tree over the original problem. (The DVO and DSO AND/OR search schemes presented by Mateescu and Dechter (2005) are also versions of DDP run with particular variable ordering heuristics). In comparison with the algorithms we present in the next section, Bayardo and Pehoushek (2000) did not provide a complexity analysis of DDP, DDP did not use caching to enhance its performance, and DDP still has less flexibility in its variable ordering. In particular, once the problem has been split into independent components the search must solve these components sequentially in separate recursions. Inside each recursion the search can only branch on the variables of the current component. That is, DDP cannot interleave the solution of these components like the DPLL algorithms we present here.

## 2.5 Complexity Analysis

All known algorithms for BAYES, #SAT and SUMPROD run in exponential-time in the worst case. However, when the branch width of the underlying hypergraph of the instance, $w$, is small, the some of the above algorithms are much more efficient. It can be shown that the algorithms VE, RC-Cache and AND/OR-Cache discussed above run in time and space $n^{O(1)}2^{O(w)}$. We note that the complexity of these algorithms is usually given in terms of tree width or elimination width, and not branch width. However, by Lemmas 1, 2, and 3, these concepts are equivalent to within a factor of 2, and therefore the asymptotic complexity can equivalently be stated in terms of any of these three notions of width (tree width, branch width, or elimination width). For analyzing our backtracking algorithms, branch width is be somewhat more natural, and for this reason we have chosen to state all complexity results in terms of branch width.

The runtime of the variable elimination algorithm is easily seen to be at most $n^{O(1)}2^{O(w)}$. To see this, notice that the algorithm proceeds in $n$ stages, removing one variable at each stage. Suppose that the algorithm is run on some variable ordering that has elimination width $v$. The algorithm removes the $i^{th}$ variable during the $i^{th}$ stage. At the $i^{th}$ stage, all functions involving this variable are merged to obtain a new function. As indicated in Section 2.4.1, computing the new function involves iterating overall possible instantiations of its variables. The runtime of this stage is therefore exponential in the number of underlying variables of the new function, which is bounded by $v$. Thus, the runtime of the algorithm is bounded by $n^{O(1)}2^{O(v)}$. Now by Lemmas 1 and 2 and 3, if the elimination width is $v$, then the branch width is at most $v + 1$, and therefore the overall runtime is as claimed. It can also be noted that since the new function must be stored, the space complexity of variable elimination is the same as its time complexity, i.e., $n^{O(1)}2^{O(w)}$.

It has also been shown that the run times of RC-Cache and RC-Cache$^+$ are bounded by $n^{O(1)}2^{O(w)}$ (Darwiche, 2001). Further, there is a nice time-space tradeoff. That is, the space-efficient implementation of RC, RC-Space, runs in time $2^{O(w\log n)}$ but needs only space linear in the size of the input, where as RC-Cache has space complexity equal to its time complexity, $n^{O(1)}2^{O(w)}$. We will present proofs showing that our DPLL based algorithms can achieve the same time and time/space bounds; our proofs give the bounds for RC-Space, RC-Cache, and RC-Cache$^+$ as special cases.

Finally, it has been shown that AND/OR-Space runs in time $2^{O(w\log n)}$ (Dechter & Mateescu, 2007). Specifically, Dechter and Mateescu show that AND/OR-Space runs in time exponential in the height of its inputed pseudo tree, and Bayardo and Miranker (1995) show that this height is bounded $w\log n$. Lemma 1 then shows that the bound also holds for branch width. Similarly, Dechter and Mateescu (2007) show that AND/OR-Cache runs in time and space bounded by $n^{O(1)}2^{O(w)}$ by exploiting the very close relationship between pseudo trees and elimination orders.

**Making the algorithms deterministic.**    As stated above, all of these algorithms are in fact nondeterministic algorithms each requiring a different nondeterministically determined input. Hence, the stated complexity bounds mean that there exists some choice of nondeterministic input (i.e., some variable ordering for VE, some branch decomposition for RC, and some pseudo tree for AND/OR search) with which the algorithm can achieve the stated complexity bound.

However, to achieve this runtime in practice, we will need to be able to *find* such a good branch decomposition (variable ordering, pseudo tree) efficiently. Unfortunately, the general problem of computing an optimal branch decomposition (i.e., one that has width equal to the branch width of $\mathcal{H}$) is NP-complete. However, Robertson and Seymour (1995) present an algorithm for computing a branch decomposition with branch width that is within a factor of 2 of optimal and that runs in time $n^{O(1)}2^{O(w)}$, where $w$ is the branch width of $\mathcal{H}$. By first running this deterministic algorithm to compute a good branch decomposition, one can obtain *deterministic* versions of RC-Cache and RC-Cache$^+$ that run in time and space $n^{O(1)}2^{O(w)}$, as well as a *deterministic* version of RC-Space that runs in linear space and time $2^{O(w\log n)}$. These deterministic versions no longer require access to a nondeterministically determined choice to achieve their stated runtimes.

---

**Algorithm 5**: DPLL for SAT

**1 DPLL** $(\phi)$
**2 begin**
**3**    **if** $\phi$ *has no clauses* **then**
**4**        **return** TRUE
**5**    **else if** $\phi$ *contains an empty clause* **then**
**6**        **return** FALSE
**7**    **else**
**8**        **choose** a variable $x$ that appears in $\phi$
**9**        **return** ($\mathbf{DPLL}(\phi|_{x=0}) \vee \mathbf{DPLL}(\phi|_{x=1})$)
**10 end**

---

Similarly with a nearly optimal branch decomposition, we can use Lemmas 1-3 to find a nearly optimal elimination ordering, and thus can obtain a *deterministic* version of the variable elimination algorithm that runs in time and space $n^{O(1)}2^{O(w)}$. And finally, from that nearly optimal elimination ordering the bucket-tree construction of Dechter and Mateescu (2007) can be used to construct a nearly optimal pseudo tree, and thus we can obtain a *deterministic* version of AND/OR-Space that runs in linear space and time $2^{O(w \log n)}$, and a *deterministic* version of AND/OR-Cache that runs in time and space $n^{O(1)}2^{O(w)}$.

## 3. Using DPLL for #SAT and SUMPROD

Now we present our methods for augmenting backtracking search with different caching schemes so that it can solve SUMPROD with time and space guarantees at least as good as the other exact algorithm for SUMPROD. For ease in presentation we present DPLL-based algorithms for solving #SAT, and derive complexity results for these algorithms. Later we will discuss how the algorithms and complexity results can be applied to other instances of SUMPROD (like BAYES).

### 3.1 DPLL and #DPLL:

DPLL is a nondeterministic algorithm for SAT, that has also been used to solve various generalizations of SAT, including #SAT (Dubois, 1991; Zhang, 1996; Birnbaum & Lozinskii, 1999; Littman, Majercik, & Pitassi, 2001). DPLL solves SAT by performing a depth-first search in the space of partial instantiations (i.e., it is a standard backtracking search algorithm). The nondeterministic part of the computation is lies in the choice of which variable to query (i.e., instantiate) next during its search. It operates on SAT problems encoded in clause form (CNF).

The standard DPLL algorithm for solving SAT is given in Algorithm 5. We use the notation $\phi|_{x=0}$ or $\phi|_{x=1}$ to denote the new CNF formula obtained from reducing $\phi$ by setting the variable $x$ to 0 or 1. Reducing $\phi$ by $x = 1$ ($x = 0$) involves removing from $\phi$ all clauses containing $x$ ($\neg x$) and removing the falsified $\neg x$ ($x$) from all remaining clauses.

DPLL is a nondeterministic procedure that generates a *decision tree* representing the underlying CNF formula. For solving SAT, the decision tree is traversed in a depth-first manner until either a satisfying path is encountered, or until the whole tree is traversed (and all paths falsify the formula). The nondeterminism of the algorithm occurs in the choice of variable on line 8. In practice this

---

**Algorithm 6**: #DPLL for #SAT (no caching)

1  **#DPLL** $(\phi)$
   // Returns the probability of $\phi$
2  **begin**
3    **if** $\phi$ *has no clauses* **then**
4       **return** $1$
5    **else if** $\phi$ *contains an empty clause* **then**
6       **return** $0$
7    **else**
8       **choose** a variable $x$ that appears in $\phi$
9       **return** $(\frac{1}{2}\textbf{\#DPLL}(\phi|_{x=0}) + \frac{1}{2}\textbf{\#DPLL}(\phi|_{x=1}))$
10 **end**

---

nondeterminism is typically resolved via some heuristic choice. Also, the algorithm utilizes early termination of the disjunctive test on line 9; i.e., if the first test returns TRUE the second recursive call is not made. Thus, the algorithm stops on finding the first satisfying path.

Note that we do not require that DPLL perform *unit propagation*. In particular, unit propagation can always be realized through the choice of variable at line 8. In particular, if we force DPLL to always chose a variable that appears in a unit clause of $\phi$ whenever one exists, this will have the same effect as forcing DPLL to perform unit propagation after every variable instantiation. That is, after a variable is chosen, and instantiated to one of its values, the input CNF $\phi$ will be reduced. The reduced formula, $\phi|_{x=0}$ or $\phi|_{x=1}$, passed to the next recursive call may contain unit clauses. With unit propagation, the variables in these clauses would be instantiated so as to satisfy the unit clauses. If instead, we force one of these variable to be chosen next, one instantiation would immediately fail due to the generation of an empty clause, while the other would instantiate the variable to the same value as unit propagation. Hence, since we analyze DPLL as a nondeterministic algorithm, this includes those deterministic realizations that perform unit propagation.

A simple modification of DPLL allows it to count all satisfying assignments. Algorithm 6 gives the #DPLL algorithm for counting. The algorithm actually computes the probability of the set of satisfying assignments under the uniform distribution. Hence, the number of satisfying assignments can be obtained by multiplying this probability by $2^n$, where $n$ is the number of variables in $\phi$. The alternative would be to return 2 raised to the number of unset variables whenever $\phi$ has no clauses (line 4) and not multiply the recursively computed counts by $\frac{1}{2}$ (line 9).

Known exponential worst-case time bounds for DPLL also apply to #DPLL: for unsatisfiable formulas, both algorithms have to traverse an entire decision tree before terminating. Although this decision tree can be small (e.g., when an immediate contradiction is detected), for some families of formulas the decision tree must be large. In particular, it is implicit in the results of Haken (1985) that *any* decision tree for the formulas encoding the (negation of the) propositional pigeonhole principle has exponential size, and thus DPLL and #DPLL must take exponential-time on these examples. This lower bound does not, however, help us discriminate between algorithms since *all* known algorithms for #SAT and BAYES take exponential-time in the worst-case. Nevertheless, #DPLL requires exponential time even on instances that can be efficiently solved by competing algorithms for SUMPROD. To see this, consider a 3CNF formula over $3n$ variables consisting of $n$ clauses that share no variables. Any complete decision tree has exponential size, and therefore #DPLL will require exponential time. In contrast, since this formula has low tree width it can be solved in polynomial time by VE, RC, or AND/OR search.

### 3.2 DPLL with Caching:

Given that the obvious application of DPLL to solve SUMPROD can give exponentially worse performance than the standard algorithms, we now examine ways of modifying DPLL so that it can solve #SAT (and thus BAYES and SUMPROD) more efficiently. To understand the source of #DPLL's inefficiency consider the following example.

**Example 6** The following diagram shows a run of #DPLL on $\phi = \{(w \vee x)(y \vee z)\}$. Each node shows the variable to be branched on, and the current formula #DPLL is working on. The left hand branches correspond to setting the branch variable to FALSE, while on the right the variable is set to TRUE. The empty formula is indicated by $\{\}$, while a formula containing the empty clause is indicated by $\{()\}$. The diagram shows that #DPLL encounters and solves the subproblem $\{(y \vee z)\}$ twice: once along the path $(w = 0, x = 1)$ and again along the path $(w = 1)$. Note that in this example unit propagation is realized by the choice of variable ordering—after $w$ is set to FALSE, #DPLL chooses to instantiate the variable $x$ since that variable appears in a unit clause.

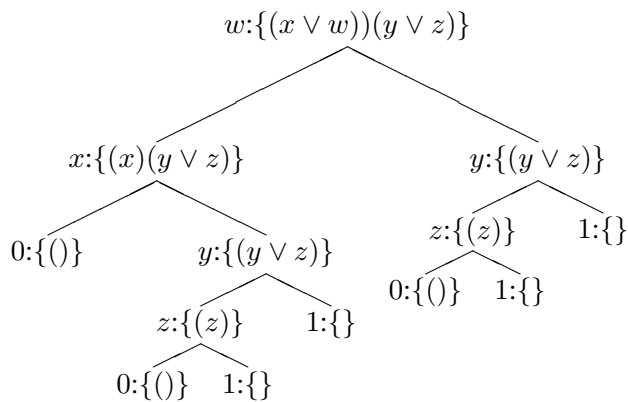

If one considers the above example of applying #DPLL to disjoint sets of clauses, it becomes clear that in some formulas #DPLL can encounter the same subproblem an exponential number of times.

### 3.2.1 DPLL with Simple Caching (#DPLL-SimpleCache)

One way to prevent this duplication is to apply memoization. As indicated in Example 6, associated with every node in the DPLL tree is a formula $f$ such that the subtree rooted at this node is trying to compute the number of satisfying assignments to $f$. When performing a depth-first search of the tree, we can keep a cache that contains all formulas $f$ that have already been solved, and upon reaching a new node of the tree we can avoid traversing its subtree if the value of its corresponding formula is already stored in the cache.

In Example 6 we would cache $\{(y \vee z)\}$, when we solve it along the path $(w = 0, x = 1)$ thereby avoid traversing the subtree below $(w = 1)$.

---

**Algorithm 7**: #DPLL algorithm with simple caching (#DPLL-SimpleCache)

---

1 **#DPLL-SimpleCache** ($\phi$)
   `// Returns the probability of` $\phi$
2 **begin**
3    **if** *InCache($\phi$)* **then**
      `// Also detects obvious formulas.`
4       **return** GetValue($\phi$)
5    **else**
6       **choose** a variable $x$ that appears in $\phi$
7       $val = \frac{1}{2}$**#DPLL-SimpleCache** ($\phi|_{x=0}$) $+ \frac{1}{2}$**#DPLL-SimpleCache** ($\phi|_{x=1}$)
8       AddToCache($\phi$,*val*)
9       **return** *val*
10 **end**

---

The above form of caching, which we will call *simple caching* (#DPLL-SimpleCache) can be easily implemented as shown in Algorithm 7.[4] As with #DPLL, #DPLL-SimpleCache returns the probability of its input formula $\phi$; multiplying this by $2^n$ gives the number of satisfying assignments.

In addition to formulas stored in the cache there are also the following **obvious** formulas whose value is easy to compute. (1) The empty formula $\{\}$ containing no clauses has value 1. (2) Any formula containing the empty clause has value 0. Obvious formulas can be treated as if they are implicitly stored in the cache (they need not be explicitly stored in the cache, rather their values can be computed as required).

The following (low complexity) subroutines are used to access the cache. (1) AddToCache($\phi, r$): adds to the cache the fact that formula $\phi$ has value $r$. (2) InCache($\phi$): takes as input a formula $\phi$ and returns true if $\phi$ is in the cache. (3) GetValue($\phi$): takes as input a formula $\phi$ known to be in the cache and returns its stored value. There are various ways of computing a cache key from $\phi$. For example, $\phi$ can be maintained as a sorted set of sorted clauses, and then cached as if it was a text string. Such a caching scheme has $n^{O(1)}$ complexity.

Surprisingly, simple caching, does reasonably well. The following theorem shows that simple caching achieves runtime bounded by $2^{O(w \log n)}$, where $w$ is the underlying branch width. As with our complexity analysis of earlier algorithms presented in Section 2.5, the simple caching algorithm can also be made *deterministic* by first computing a branch decomposition that is within a factor of 2 of optimal (using the Robertson-Seymour algorithm), and then running #DPLL-SimpleCache with a variable ordering determined by this branch decomposition.

**Theorem 1** For solving #SAT with $n$ variables, there is an execution of #DPLL-SimpleCache that runs in time bounded by $2^{O(w \log n)}$ where $w$ is the underlying branch width of the instance. Furthermore, the algorithm can be made deterministic with the same time guarantees.

Although the theorem shows that #DPLL-SimpleCache does fairly well, its performance is not quite as good as the best SUMPROD algorithms (which run in time $n^{O(1)} 2^{O(w)}$).

---

4. Simple caching has been utilized before (Majercik & Littman, 1998), but without theoretical analysis.

---

**Algorithm 8**: #DPLL algorithm with component caching (#DPLL-Cache)

---

1   **#DPLL-Cache** ($\Phi$)
     // Returns the probability of the set of disjoint formulas $\Phi$
2   **begin**
3      **if** *InCache*($\Phi$) **then**
         // Also detects obvious formulas.
4          **return** GetValue($\Phi$)
5      **else**
6          $\Psi$ = RemoveCachedComponents($\Phi$)
7          **choose** a variable $x$ that appears in some component $\phi \in \Psi$
8          $\Psi^- =$ ToComponents($\phi|_{v=0}$)
9          **#DPLL-Cache** ($\Psi - \{\phi\} \cup \Psi^-$)
10        $\Psi^+ =$ ToComponents($\phi|_{v=1}$)
11        **#DPLL-Cache** ($\Psi - \{\phi\} \cup \Psi^+$)
12        AddToCache($\phi, \frac{1}{2}$GetValue($\Psi^-$) $+ \frac{1}{2}$GetValue($\Psi^+$))
13        **if *#DPLL-Space*** **then**
14          RemoveFromCache($\Psi^- \cup \Psi^+$)
15        **return** GetValue($\Phi$)
16   **end**

---

### 3.2.2 DPLL with Component Caching (#DPLL-Cache)

Now we show that a more sophisticated caching scheme allows #DPLL to perform as well as the best known algorithms. We call the new algorithm #DPLL-Cache, and its implementation is given in Algorithm 8.

In the algorithm we generalize the cache to deal with sets of formulas. First, we say that a (single) formula $\phi$ is *known* if its value is stored in the cache or it is an obvious formula (and its value is implicitly stored in the cache). Given a set of formulas $\Phi$ we say that the set is *known* if either every $\phi \in \Phi$ is known, or there is some $\phi \in \Phi$ whose value is known to be zero. In both cases we say that $\Phi$'s value is equal to the product of the values of the $\phi \in \Phi$.

Now we generalize some of the cache access subroutines. (1) InCache($\Phi$) is generalized so that it can take as input a set of formulas $\Phi$. It returns true if $\Phi$ is known as just defined. (2) Similarly GetValue($\Phi$) is generalized to take sets of formulas as input. It returns the product of the cached values of the formulas $\phi \in \Phi$.

The intuition behind #DPLL-Cache is to recognize that as variables are set the input formula may become broken up into disjoint components, i.e., sets of clauses that share no variables with each other. Since these components share no variables we can compute the number of solutions to each component and multiply the answers to obtain the total solution count. Thus, it is intended that GetValue be called with a set of disjoint components $\Phi$. In that case it will correctly return the solution count for $\Phi$—i.e., the product of the solution counts for each $\phi \in \Phi$.

The algorithm creates a standard DPLL tree, however it caches component formulas as their values are computed. It keeps its input in decomposed form as a set of disjoint components, and if any of these components are already in the cache (and thus their value is known) it can remove these parts of the input—reducing the size of the problem it still has to solve and avoiding having to resolve these components.

The new algorithm uses the previously defined cache access subroutines along with two additional (low complexity) subroutines. (1) ToComponents($\phi$): takes as input a formula $\phi$, breaks it up into a set of minimal sized disjoint components, and returns this set. (2) RemoveCachedComponents($\Phi$): returns the input set of formulas $\Phi$ with all known formulas removed. The input to #DPLL-Cache is always set of disjoint formulas. Hence, to run #DPLL-Cache on the input formula $\phi$ we initially make the call #DPLL-Cache (ToComponents($\phi$)).

ToComponents simply computes the connected components of the primal graph generated by $\phi$. That is, in this graph all of the variables of $\phi$ are nodes, and two nodes are connected if and only if the corresponding variables appear together (in any polarity) in a clause of $\phi$. Each connected component of this primal graph (which can be computed with a simple depth-first traversal of the graph Cormen, Leiserson, Rivest, & Stein, 2001), defines a set of variables whose clauses form an independent component of $\phi$.

Each call of #DPLL-Cache completes with the solution of the unknown components from the set of inputed components $\Phi$. If all components of $\Phi$ are known the product of the values of these components will be returned at line 4. Otherwise the input set of components is reduced by removing all known components (line 6), which must leave at least one unknown component and potentially reduces the size of the remaining problem to be solved. Then a variable from some unsolved component is chosen and is branched on. Since the variable only appears in the component $\phi$ its assignment can only affect $\phi$. In particular, its assignment might break $\phi$ into smaller components (line 8 and 11). The recursive call will solve all components it is passed, so after the two recursive calls the value of $\phi$ can be computed and cached (line 12). Finally, since all components in the inputed set $\Phi$ are now solved its value can be retrieved from the cache and returned.

**Example 7** Figure 5 illustrates the behavior of #DPLL-Cache on the formula $\phi = \{(a, b, c, x), (\neg a, b, c), (a, \neg b, c), (d, e, f, x), (\neg d, e, f), (d, \neg e, f)\}$. Although the problem could be solved with a simpler search tree, we use a variable ordering that generates a more interesting behavior.

Each node shows the variable to be branched on, and the current set of components #DPLL-Cache is working on. The known components (i.e., those already in the cache) are marked with an asterisk (*). The branch variables are set to FALSE on the left branch and TRUE on the right branch. The empty formula is indicated by $\{\}$, while a formula containing the empty clause is indicated by $\{()\}$. To simply the diagram we use unit propagation to simplify the formula after the branch variable is set. This avoids the insertion into the diagram of nodes where unit clause variables are branched on. Finally, note that known formulas are removed before a recursive call is made, as per line 6 of Algorithm 8).

At the root, once $x$ has been set to false, $\phi$ is broken up into two components $\phi_{a,b,c} = \{(a, b, c), (\neg a, b, c), (a, \neg b, c)\}$, and $\phi_{d,e,f} = \{(d, e, f), (\neg d, e, f), (d, \neg e, f)\}$. The search tree demonstrates that it does not matter how the search interleaves branching on variables from different components, the components will still be solved independently. We see that the leftmost node in the tree that branches on $f$ succeeds in solving the component $\{(e, f), (\neg e, f)\}$. This component is then added to the cache. Similarly, the parent node that branches on $b$ solves the component $\{(b, c), (\neg b, c)\}$. (The subcomponents $\Psi^-$ and $\Psi^+$ generated by setting $b$, lines 8 and 11 of Algorithm 8, and performing unit propagation are equal to the empty formula, $\{\}$, and thus are known). On backtrack to $d$, the alternate value for $d$ does not affect the component $\{(b, c), (\neg b, c)\}$, so its value can be retrieved from the cache leaving only the component $\{(e, f)\}$ to be solved. Branching on $e$ solves this component. Backtracking to $d$ we have that both $\{(e, f)\}$ and $\{(e, f), (\neg e, f)\}$ are solved, so $\phi_{d,e,f}$'s value can be computed and placed in the cache. On backtracking to $a$, the alternate value for $a$ does not affect the component $\phi_{d,e,f}$, so its value can be retrieved from the cache leaving only the component $\{(b, c)\}$ to be solved. Branching on $b$ solves this component, after which both $\{(b, c)\}$ and $\{(b, c), (\neg b, c)\}$ are solved so $\phi_{a,b,c}$'s value can be computed and placed in the cache. The search can then backtrack to try setting $x$ to TRUE.

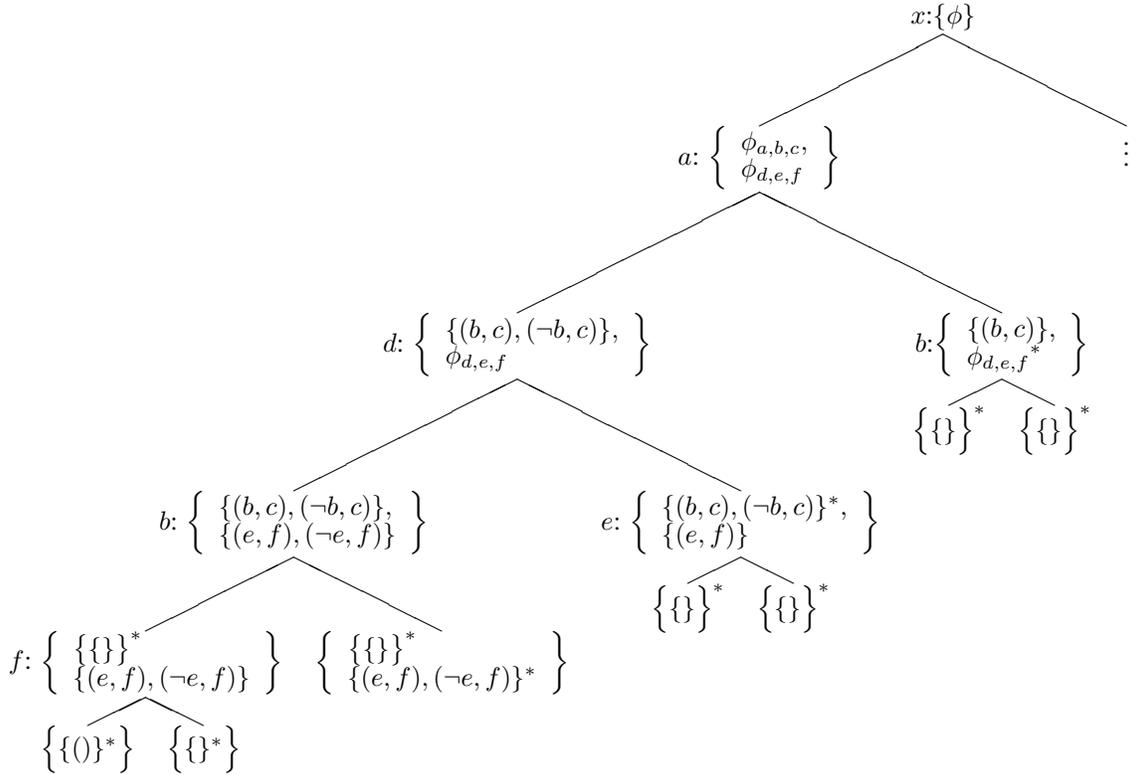

Figure 5: Search Space of #DPLL-Cache

We can obtain the following upper bound on the runtime of #DPLL-Cache.

**Theorem 2** For solving #SAT on $n$ variables, there exists an execution of #DPLL-Cache that runs in time bounded by $n^{O(1)}2^{O(w)}$ where $w$ is the underlying branch width of the instance. Furthermore, the algorithm can be made deterministic with the same time guarantees (as discussed in Section 2.5).

So we see that #DPLL-Cache can achieve the same level of performance as the best SUMPROD algorithms.

Finally, there is a third variant of #DPLL with caching, **#DPLL-Space**, that achieves a nontrivial time-space tradeoff. This algorithm is the natural variant of #DPLL-Cache, modified to remove cached values so that only linear space is consumed. The algorithm utilizes one additional subroutine. (6) RemoveFromCache($\Phi$): takes as input a set of formulas (a set of components) and removes all of them from the cache. After splitting a component with a variable instantiation and computing the value of each part, #DPLL-Space cleans up the cache by removing all of these sub-components, so that only the value of the whole component is retained. Specifically, #DPLL-Space is exactly like #DPLL-Cache, except that it calls RemoveFromCache($\Psi^{-} \cup \Psi^{+}$) just before returning (line 14).

**Theorem 3** For solving #SAT on $n$ variables, there is an execution of #DPLL-Space that uses only space linear in the instance size and runs in time bounded by $2^{O(w \log n)}$ where $w$ is the underlying branch width of the instance. Furthermore, the algorithm can be made deterministic with the same time and space guarantees.

The proofs of Theorems 1–3 are given in the appendix.

### 3.3 Using DPLL Algorithms for Other Instances of SUMPROD:

The DPLL algorithms described in this section can be easily modified to solve other instances of SUMPROD. However, since #SAT is #P complete many instances of SUMPROD can also be solved by simply encoding them in #SAT. For example, this approach is readily applicable to BAYES and has proved to be empirically successful (Sang et al., 2005b). Furthermore, the encoding provided by Sang et al. (2005b) achieves the same complexity guarantees as standard algorithms for BAYES. (That is, the CNF encoding has tree width no greater than the original Bayes Net). Note that this encoding assigns non-uniform probabilities to values of the variables. That is, for variable $x$ the probability of $x = 0$ might not be equal to the probability of $x = 1$. This is easily accommodated in our algorithms: instead of multiplying the value returned by each recursive call by $\frac{1}{2}$ we simply multiply it by the probability of the corresponding variable value (i.e., by $Pr(x = 0)$ or $Pr(x = 1)$).

On the other hand, if conversion to #SAT is inapplicable or undesirable the algorithms can be modified to solve other instances of SUMPROD directly. For SUMPROD, we want to compute $\bigoplus_{X_1} \cdots \bigoplus_{X_n} \bigotimes_{j=1}^{m} f_j(E_j)$. DPLL chooses a variable, $X_i$, and for each value $d$ of $X_i$ it recursively solves the reduced problem $\mathcal{F}|_{X_i=d}$. (Hence, instead of a binary decision tree it builds a $k$-ary tree). The reduced problem $\mathcal{F}|_{X_i=d}$ is to compute

$$\bigoplus_{X_1} \cdots \bigoplus_{X_{i-1}} \bigoplus_{X_{i+1}} \cdots \bigoplus_{X_m} \bigotimes_{j=1}^{m} f_j(E_j)|_{X_i=d},$$

where $f_j(E_j)|_{X_i=d}$ is $f_j$ reduced by setting $X_i = d$. #DPLL-SimpleCache caches the solution to the reduced problem to avoid recomputing it. For example, it can remember the reduced problem by remembering which of the original functions in $\mathcal{F}$ remain (i.e., have not been reduced to a constant value) and the set of assignments that reduced these remaining functions. #DPLL-Cache caches the solution to components of the reduced problem. For example, it can remember a component by remembering the set of original functions that form the component along with the set of assignments that reduced these functions. It can compute the current components by finding the connected components of the primal graph generated from the hypergraph of the SUMPROD instance with all instantiated variables removed. It is a straightforward adaptation to show that the above three theorems continue to hold for #DPLL, #DPLL-Cache, and #DPLL-Space so modified to solve SUMPROD.

Algorithm 9 shows how #DPLL-Cache, for example, can be modified to solve general SUMPROD problems. The algorithm takes as input a set of components $\Phi$, just like #DPLL-Cache, initially containing the components of the original problem. In the algorithm $fns(x)$ denotes the set of functions of the original problem that (a) contain $x$ in their scope, and (b) are fully instantiated by the instantiation of $x$.

---

**Algorithm 9**: SUMPROD-DPLL-Cache algorithm for arbitrary SUMPROD problems

---

**1** **SUMPROD-DPLL-Cache** ($\Phi$)

**2** **begin**

**3**     **if** *InCache*($\Phi$) **then**

**4**         **return** GetValue($\Phi$)

**5**     **else**

**6**         $\Psi$ = RemoveCachedComponents($\Phi$)

**7**         **choose** a variable $x$ that appears in some component $\phi \in \Psi$

**8**         $p = 0$

**9**         **foreach** $d \in$ *domain of $x$* **do**

**10**             $\Phi^d$ = ToComponents($\phi|_{x=d}$)

**11**             $\alpha = \prod_{f \in fns(x)}$ LOOKUP(value of $f$ on $\rho \cup \{x = d\}$)

**12**             $p = p + \alpha \times$ **SUMPROD-DPLL-Cache**($\Phi - \{\phi\} \cup \Phi^d$)

**13**         **end**

**14**         AddToCache($\phi, p$)

**15**     **return** GetValue($\Phi$)

**16** **end**

---

## 4. Comparing Algorithms for BAYES and #SAT

In this section, we will prove that our DPLL based algorithms are at least as powerful as the standard complete algorithms for solving #SAT, and that they are provable more powerful than many of them on some instances. This last feature is important as it means that solving SUMPROD using DPLL augmented with caching can in some cases solve problems that are beyond the reach of many standard complete algorithms.

As mentioned earlier, the algorithms for SUMPROD as well as our new DPLL-based algorithms, are actually *nondeterministic* algorithms that require some nondeterministically chosen input. (This input can be viewed as being a sequence of bits). For VE, the nondeterministic bits encode an elimination ordering; for RC, the nondeterministic bits encode a branch decomposition; for AND/OR search the nondeterministic bits encode a pseudo tree; and for our DPLL based algorithms, the nondeterministic bits encode the underlying decision tree indicating which variable will be queried next in the backtracking process. Thus when comparing the "power" of these algorithms we must be careful about how the nondeterminism is resolved. For example, VE operating with a very bad elimination ordering cannot be expected to run as efficiently as #DPLL-Cache operating with a very good branching strategy. First we present some definitions which allow us to state our results precisely.

**Definition 7** Let $f$ be a CNF formula. Define $Time[\text{VE}](f)$ to be the minimal runtime of any variable elimination algorithm for solving #SAT for $f$, over all choices of elimination orderings for $f$. Similarly define $Time[A](f)$, for $A$ equal to RC-Cache, RC-Space, RC-Cache$^+$, AND/OR-Space, AND/OR-Cache, AND/OR-Cache$^+$, #DPLL-Cache, and #DPLL-Space. (For example, $Time[\text{RC-Cache}](f)$ is the minimal runtime of the RC-Cache algorithm solving #SAT for $f$, over all possible branch decompositions of $f$.)

**Definition 8** Let $A$ and $B$ be two nondeterministic algorithms for #Sat. Then we will say that $A$ polynomial-time simulates $B$ if there is a fixed polynomial $p$ such that for every CNF formula $f$ $Time[A](f) \leq p(Time[B](f))$.

The following theorem shows that RC-Cache and RC-Cache$^+$ polynomially simulate VE. The proof of this theorem is implicit in the results of Darwiche (2001).

**Theorem 4** Both RC-Cache and RC-Cache$^+$ polynomially simulate VE.

Now we prove that DPLL with caching is as powerful as previous algorithms.

**Theorem 5** #DPLL-Cache polynomially simulates RC-Cache, RC-Cache$^+$, AND/OR-Cache, AND/OR-Cache$^+$, and VE. #DPLL-Space polynomially simulates RC-Space, AND/OR-Space and DDP.[5]

The proof of this theorem is given in the appendix. It should be noted that the proof also implies that there is a deterministic version of #DPLL-Cache that has time (and space) complexity that is at least as good as any deterministic realization of RC-Cache, RC-Cache$^+$, AND/OR-Cache, AND/OR-Cache$^+$, or VE. Similarly, there is a deterministic version of #DPLL-Space that has time (and space) complexity that is at least as good as any deterministic realization of RC-Space, AND/OR-Space and DDP.

Now we prove that DPLL with caching can in some cases run super-polynomially faster than previous algorithms. The proof is given in the appendix.

**Theorem 6** None of RC-Space, RC-Cache, AND/OR-Cache, AND/OR-Space or VE can polynomially simulate #DPLL-Cache, #DPLL-Space, or #DPLL.

This theorem shows that #DPLL-Cache/Space has a basic advantage over the other standard algorithms for SumProd. That is, on some problems RC, AND/OR search, and VE will all require time super-polynomially greater than #DPLL-Cache no matter what branch decomposition, pseudo tree, or variable ordering they are supplied with, even when caching is utilized. The proof of this theorem shows that the advantage of #DPLL-Cache arises from its ability to utilize dynamic variable orderings, where each branch can order the variables differently. The flexibility of a dynamic variable ordering for these instances gives rise to increased opportunities for contradictions thereby significantly decreasing the overall runtime.

We note that Theorem 6 does not cover those algorithms that have more flexibility in their variable ordering, i.e., AND/OR-Cache$^+$, RC-Cache$^+$, and DDP. It is an open problem whether or not #DPLL-Cache is superpolynomially faster than these algorithms on some instances, although we conjecture that Theorem 6 is also true for these algorithms.

In particular, note that #DPLL-Cache still has greater flexibility in its variable ordering than any of these algorithms. None of these algorithms have complete flexibility in their variable ordering. AND/OR-Cache$^+$ must select an uninstantiated variable from the chain that starts at the root of its passed pseudo tree; RC-Cache$^+$ must select an uninstantiated variable from the intersection of the labels of the left and right children of the root of its passed branch decomposition; and DDP must select an uninstantiated variable from the component it is currently solving. In contrast #DPLL-Cache can select any uninstantiated variable.

---

5. DDP is the algorithm presented by Bayardo and Pehoushek (2000).

The difficulty with proving Theorem 6 for these other algorithms is that all of them can trade-off flexibility in their variable ordering with their ability to decompose the problem. The clearest example of this occurs with AND/OR-Cache$^+$. If AND/OR-Cache$^+$ is passed a pseudo tree that is simply a single chain of variables, it will have complete flexibility in its variable ordering, but at the same time it will never decompose the problem. Similarly, if RC-Cache$^+$ is provided with a branch decomposition that has large labels it will have more flexibility in its variable ordering, but will be less effective in decomposing the problem. For the family of problems used to prove Theorem 6 only flexibility in the variable ordering is needed to achieve a superpolynomial speedup, and thus for example AND/OR-Cache$^+$ can achieve this speedup by completely sacrificing decomposition.

#DPLL-Cache can manage the tradeoff between flexibility in variable ordering and decomposing the problem in more sophisticated ways. For example, it has the ability to use a variable ordering that encourages decomposition in some parts of its search tree while using a different variable orderings in other parts of its search tree. For instance, the Cachet system, which is based on #DPLL-Cache, employs a heuristic that dynamically trades off a variable's ability to decompose the problem with its ability to refute the current subtree (Sang et al., 2005a). (It employs a weighted average of the number of clauses the variable will satisfy and the variable's VSID score Moskewicz, Madigan, Zhao, Zhang, & Malik, 2001). #DPLL-Cache also has the ability to interleave the solving of its current set of components by successively choosing variables from different components. To extend Theorem 6 to cover AND/OR-Cache$^+$, RC-Cache$^+$ and DDP a family of problems exploiting these features of #DPLL-Cache would have to be developed.

## 5. Impact on Practice

Some of the results of this paper were first presented in a conference paper (Bacchus, Dalmao, & Pitassi, 2003), and since that time a number of works have been influenced by the algorithmic ideas presented here.

The Cachet system (Sang et al., 2004, 2005a) is a state of the art #SAT solver directly based on the results presented here. Cachet like our #DPLL-Cache algorithm, is based on the ideas of dynamic decomposition into components and caching of component solutions. It was an advance over previous #SAT solvers in its use of caching to remember previously solved components and in its integration of clause learning. The previous best #SAT solver, the DDP solver (Bayardo & Pehoushek, 2000), also performed dynamic component detection but had neither component caching nor clause learning. Our results highlighted the importance of component caching and the possibility of basing a #SAT solver on a standard DPLL implementation thus making the integration of clause learning feasible.

Cachet resolved a number of issues in making the algorithms we presented here practical. This included practical ways of implementing the caching of components including a method for efficiently computing a key that could be used for cache lookup. (This method was subsequently improved by Thurley, 2006). The Cachet system has also been used to solve BAYES, most probable explanation (MPE), and weighted MAX-SAT problems by encoding these problems as weighted #SAT problems (Sang et al., 2005b, 2007). This approach has proved to be very successful, especially for BAYES where it is often much superior to standard BAYES algorithms. The applications of #SAT and the Cachet system for BAYES has been further advanced by Li et al. (2006, 2008).

It should also be noted that practical #SAT solving and its applications to other problems like BAYES has also been advanced during this period by work on the RC algorithm and its application to compiling CNF into representations on which model counting is tractable, e.g., (Darwiche, 2004; Chavira & Darwiche, 2006, 2008). This work has also illustrated the value of converting various problems into weighted #SAT instances, and the utilization of techniques like clause learning (in this case integrated into a RC style algorithm). There has also been considerable work advancing AND/OR search, e.g., (Dechter & Mateescu, 2004; Marinescu & Dechter, 2006; Dechter & Mateescu, 2007).

One difference between the Cachet system and the RC and AND/OR search based systems mentioned above is that Cachet utilized a dynamic decomposition scheme. In particular, Cachet used a dynamic variable ordering heuristic that attempts to trade off a variable's ability to decompose the problem with its ability to refute the current subtree. Because the variable ordering was dynamically determined during search, Cachet cannot predict what components will be generated during search. Hence it has to examine the current component (i.e., the component containing the variable just instantiated) to discover the new components generated. Thus Cachet utilized an approach like that specified in Algorithm 8 where a function like ToComponents is invoked on newly reduced component (see line 8). ToComponents must do a linear computation to find the new components (e.g., a depth-first search or a union-find algorithm). In addition, for each component it must examine the clauses contained in the component to compute a cache key.

In contrast, RC and AND/OR search take as input a static or precomputed decomposition scheme (i.e., a branch decomposition or a pseudo tree). Hence, they are able to find components without doing any extra work during search, and are able to more efficiently compute cache keys for these components. For example, with AND/OR search, the algorithm simply follows the supplied pseudo tree. When the variable $V$ along with all variables on the path from the root to $V$ have been instantiated, AND/OR search knows that the variables in each subtree rooted by a child of $V$ forms an independent component. Hence, it can "detect" these components during search in constant time. Similarly, it need not examine the clauses over the variables in these new components to compute a cache key. Instead it can compute a cache key from the node of the pseudo tree that roots the component and the set of instantiations of the parents of that root that appear in clauses with the variables of the component. Note that, the set of parents whose instantiations are relevant can be computed before search so that all that has to be done during search is to look up their current values.

Thus, by using a static decomposition scheme RC and AND/OR search can gain efficiency over Cachet. However, these statically computed decompositions are not always as effective as the dynamic scheme employed by Cachet. First, it can be useful to override the precomputed decomposition scheme so as to drive the search towards contradictions. This is the gist of Theorem 6 which shows that more dynamic flexibility in variable ordering can provide superpolynomial reductions in the size of the explored search tree by better exploiting such contradictions. Second, static decompositions cannot account for the different values of the variables. That is, the formula that arises after instantiating a variable $V$ to 0 can be quite different from the formula that arises after instantiating $V$ to 1. This difference can negatively affect the performance of RC and AND/OR search in at least a couple of ways: components might be generated that are not predicted by the static decomposition scheme and thus the static scheme might not fully exploit decomposition; and due to the specific changes to the formula generated by particular instantiations, the static decomposition scheme might be inappropriate for much of the search space.

In practice, Cachet displays a performance that is at least as good as systems built using the RC algorithm, and in some cases its performance is superior (see the empirical results presented by Sang et al., 2004, 2005a). It should also be noted that #DPLL-Cache can easily utilize a static decomposition scheme and gain all of the efficiencies of such schemes. For example, if provided with a pseudo tree #DPLL-Cache can follow any ordering of the variables in that pseudo tree under which parents are always instantiated before their children. Like AND/OR search it will know that the children of any node in the pseudo tree each root an independent component, so it also will be able to detect these components in constant time. Furthermore, it would be able to utilize the more efficient caching scheme of AND/OR search. In this case its advantage over AND/OR search would be that it would have the freedom to interleave the solving of its components.[6]

In more recent work our algorithms have also been applied to optimization problems (Kitching & Bacchus, 2008). This work involved adding branch and bound techniques to the decomposition and component caching described in #DPLL-Cache. During branch and bound dynamic variable ordering can be very effective. In particular, one wants to branch on variables that will drive the value of the current path towards a better value as this can generate a global bound that can be more effective in pruning the rest of the search space. The empirical results of Kitching and Bacchus (2008) show that the added flexibility of #DPLL-Cache can sometimes yield significant performance improvements over AND/OR search even when the extra flexibility of AND/OR-Cache$^+$ is exploited.

## 6. Final Remarks

In this paper we have studied DPLL with caching, analyzing the performance of various types of caching for #SAT. Our results apply immediately to a number of instances of the SUMPROD problem including BAYES, since #SAT is complete for the class #P. However, our proofs can also be modified without much difficulty so that our complexity results apply directly to any problem in SUMPROD.

More sophisticated caching methods have also been explored for solving SAT by Beame et al. (2003) who showed that some of these methods can considerably increase the power of DPLL. However, these more sophisticated caching methods are currently not practical due to their large overheads. In other related work, one of the results of Aleknovich and Razborov (2002) showed that SAT could be solved in time $n^{O(1)}2^{O(w)}$. Our results extend this to any problem in SUMPROD—as shown in Section 2.1 SAT is an instance of SUMPROD.

We have proved that from a theoretical point of view, #DPLL-Cache is just as efficient in terms of time and space as other state-of-the-art exact algorithms for SUMPROD. Moreover, we have shown that on specific instances, #DPLL-Cache substantially outperforms the basic versions of these other algorithms. The empirical results presented in the works described in Section 5 indicate that these advantages can often be realized in practice and that on some problems our DPLL based algorithms can yield significant performance improvements.

There are a number of reasons why our DPLL based algorithms can outperform traditional algorithms for SUMPROD. Algorithms like VE and the join tree algorithm (which is used in many BAYES inference systems), take advantage of the global structure of interconnections between the functions as characterized by the tree width or branch width of the instance. Our DPLL algorithms however, can also naturally exploit the internal or local structure within the functions. This is accomplished by instantiating variables and reducing the functions accordingly. This can lead to improvements especially when the functions are encoded in a way to expose more of the function's internal structure, such as an encoding by sets of clauses (e.g., see Li et al., 2008). There are two prominent examples of structure that can be exploited by DPLL.

---

6. Marinescu and Dechter (2007) present a method for searching an AND/OR tree in a best-first manner. This method can also interleave the solving of components, but in general best-first search has exponential space overheads.

First, some of the subproblems might contain zero valued functions. In this case our algorithms need not recurse further—the reduced subproblem must have value $0$.[7] In VE the corresponding situation occurs when one of the intermediate functions, $F_i$, produced by summing out some of the variables, has value $0$ for some setting of its inputs. In VE there is no obvious way of fully exploiting this situation. VE can achieve some gains by ignoring those parts of $F_i$'s domain that map to $0$ when $F_i$ appears in a product with other functions. However, it can still expend considerable effort computing some other intermediate function $F_j$ many of whose non-zero values might in fact be irrelevant because they will eventually be multiplied by zero values from $F_i$.

Second, it can be that some of the input functions become constant prior to all of their variables being set (e.g., a clause might become equivalent to TRUE because one of its literals has become true), or they might become independent of some of their remaining variables. This means the subproblems $f|_{x_i=1}$ and $f|_{x_i=0}$ might have quite different underlying hypergraphs. Our DPLL-based algorithms can take advantage of this fact, since they work on these reduced problems separately. For example, our algorithms are free to use dynamic variable orderings, where a different variable ordering is used solving each subproblem. VE, on the other hand, does not decompose the problem in this way, and hence cannot take advantage of this structure.

In BAYES this situation corresponds to context-specific independence where the random variable $X$ might be dependent on the set of variables $W, Y, Z$ when considering all possible assignments to these variables (so $f(X, W, Y, Z)$ is one of the input functions), but when $W = \text{True}$ it might be that $X$ becomes independent of $Y$ (i.e., $f(X, W, Y, Z)|_{W=1}$ might be a function $F(X, Z)$ rather than $F(X, Y, Z)$). Previously only ad-hoc methods have been proposed (Boutilier, Friedman, Goldszmidt, & Koller, 1996) to take advantage of this kind of structure.

It should be noted however, that when the problem's functions have little or internal structure VE can be significantly more efficient than any of the other algorithms (RC, AND/OR search and our DPLL algorithms). VE only uses simple multiplication and summation operations and does have any of the overheads involved with instantiating variables and exploring an AND/OR search tree or backtracking tree.

RC and AND/OR search share some of the same advantages over VE. However, they do not have as much flexibility as our DPLL algorithms. We have shown in Theorem 6 that fully exploiting the zero valued functions can in some instances require dynamic variable orderings that lie outside of the range of the basic versions of RC and AND/OR search. Although our proof does not cover the enhanced versions of RC and AND/OR (RC-Cache$^+$ and AND/OR-Cache$^+$), we have pointed out that even these versions do not have the same flexibility as our DPLL algorithms. In practice, the empirical evidence provided by the Cachet system (Sang et al., 2004, 2005a) and by the branch and bound system described by Kitching and Bacchus (2008) support our belief that this added flexibility can be important in practice.

The exploitation of context-specific independence also poses some problems for RC and AND/OR search algorithms. In particular, the static decomposition schemes they employ are incapable of fully exploiting this structure—as pointed out above the underlying hypergraphs of the subproblems arising from different instantiations can be radically different. However, although our DPLL algorithms are in principle able to exploit such structure, it remains an open problem to find practical ways to accomplishing this. Specifically, when a decomposition scheme is computed prior to search sophisticated (and computationally complex) algorithms can be utilized. It is difficult to overcome the overhead of such methods when they are used dynamically during search (although see Li and van Beek 2004 for some work in this direction). The development of methods that are light weight enough to use during search and are still effective for selecting decomposition promoting variables remains an open problem.

---

7. For #SAT this corresponds to the situation where a clause becomes empty.

Finally, as shown in the proof of Theorem 5, RC and AND/OR search possess no intrinsic advantages over our DPLL algorithms except perhaps conceptually simplicity. The proof shows that our DPLL algorithms can simulate RC and AND/OR search in such a way that no additional computation is required. Furthermore, as pointed out in the Section 5 our algorithms are also able to utilize static decomposition schemes obtaining the same efficiency gains as RC and AND/OR search.

Recently, several papers (Sanner & McAllester, 2005; Mateescu & Dechter, 2007) have made significant progress on developing more compact representations for functions (rather than tabular form), thereby potentially enhancing all of the algorithms discussed in this paper (VE, RC, etc.) by allowing them to exploit additional local structure within the functions. An interesting future step would be to combine the unique dynamic features of #DPLL-Cache with one of these promising compact function representations to try to further improve SUMPROD algorithms.

**Acknowledgments**   This research funded by governments of Ontario and Canada through their NSERC and PREA programs. Some of the results of this paper were presented in an earlier conference paper (Bacchus et al., 2003). We thank Michael Littman for valuable conversations.

## Appendix A. Proofs

### A.1 Lemmas Relating Branch Width, Tree Width, and Elimination Width

**Lemma 2** Let $\mathcal{H} = (V, E)$ be a hypergraph with a tree-decomposition of width $w$. Then there is an ordering $\pi$ of the vertices $V$ such that the induced width of $\mathcal{H}$ under $\pi$ is at most $w$.

**Proof:** Let $\mathcal{H} = (V, E)$ be a hypergraph of tree width $w$ and let $T_{td}$ be a tree decomposition that achieves width $w$. That is, the maximum sized label of $T_{td}$ is of size $w + 1$. We can assume without loss of generality that the labels of the leaves of $T_{td}$ are in a one-to-one correspondence with the edges of $\mathcal{H}$. For an arbitrary node $m$ in $T_{td}$, let $label(m)$ be the set of vertices in the label of $m$, $A^m$ be the tree rooted at $m$, $vertices(m)$ be the union of the labels of the leaf nodes in $A^m$ (i.e., the hyperedges of $\mathcal{H}$ appearing below $A^m$), and $depth(m)$ be the distance from $m$ to the root.

Let $x$ be any vertex of $\mathcal{H}$, and let $leaves(x)$ be the set of leaves of $T_{td}$ that contain $x$ in their label. We define $node(x)$ to be the deepest common ancestor in $T_{td}$ of all the nodes in $leaves(x)$, and the depth of a vertex, $depth(x)$, to be $depth(node(x))$. Note that $x \in label(node(x))$, since the path from the left-most leaf in $leaves(x)$ to the right-most leaf must pass through $node(x)$; and that $x$ does not appear in the label of any node outside of the subtree rooted at $node(x)$, since no leaf outside of this subtree contains $x$.

Finally let $\pi = x_1, \ldots, x_n$ be any ordering of the vertices such that if $depth(y) < depth(x)$, then $y$ must precede $x$ in the ordering. We use the notation $y <_\pi x$ to indicate that $y$ precedes $x$ in the ordering $\pi$ (and thus $y$ will be eliminated after $x$). We claim that the induced width of $\pi$ is at most the width of $T_{td}$, i.e., $w$.

Consider $A^{node(x)}$, the subtree rooted at $node(x)$, and $vertices(node(x))$, the union of the labels of the leaves of $A^{node(x)}$. We make the following observations about these vertices.

1. If $y \in vertices(node(x))$ and $y <_\pi x$, then $y$ labels $node(x)$ and $node(y)$ must be ancestor of $node(x)$ (or equal). $y <_\pi x$ implies that $depth(y) \leq depth(x)$. There must be a path from the leaf in $A^{node(x)}$ containing $y$ to $node(y)$, and since $node(y)$ is at least as high as $node(x)$ the path must go through $node(x)$ (or we must have $node(x) = node(y)$). In either case $y \in label(node(x))$.

2. If $y \in vertices(node(x))$ and $y >_\pi x$ then $node(y)$ must lie inside $A^{node(x)}$ and $node(y)$ must be a descendant of $node(x)$ (or equal). $y >_\pi x$ implies that $depth(y) \geq depth(x)$. There must be a path from the leaf in $A$ containing $y$ to $node(y)$, and since $node(y)$ is at least as deep at $node(x)$ there must either be a further path from $node(y)$ to $node(x)$, or $node(y) = node(x)$.

Note further that condition 2 implies that if $y >_\pi x$ and $y$ appears in the subtree below $node(x)$, then all hyperedges in the original hypergraph $H$ containing $y$ must also be in the subtree below $node(x)$.

We claim that the hyperedge produced at stage $i$ in the elimination process when $x_i$ is eliminated is contained in $label(node(x_i))$. Since the size of this set is bounded by $w + 1$, we thus verify that the induced width of $\pi$ is bounded by $w$ (note that the hyperedge produced in elimination does not contain $x_i$ where as $label(node(x_i))$ does).

The base case is when $x_1$ is eliminated. All hyperedges containing $x_1$ are contained in the subtree below $node(x_1)$, thus the hyperedge created when $x_1$ is eliminated is contained in $vertices(node(x_1))$. All other vertices in $vertices(node(x_1))$ follow $x_1$ in the ordering so by the above they must label $node(x_1)$ and $vertices(node(x_1)) \subseteq label(node(x_1))$.

When $x_i$ is eliminated there are two types of hyperedges that might be unioned together: (a) those hyperedges containing $x_i$ that were part of the original hypergraph $\mathcal{H}$, and (b) those hyperedges containing $x_i$ that were produced as $x_1, \ldots, x_{i-1}$ were eliminated. For the original hyperedges, all these are among the leaves below $node(x_i)$, and thus are contained in $vertices(node(x_i))$. For a new hyperedge produced by eliminating one of the previous variables, say the variable $y$, the hyperedge it produced is contained in $label(node(y))$ by induction, which in turn is contained in $vertices(node(y))$. If $y$ is in the subtree below $node(x)$ we get that this hyperedge is contained in $vertices(node(x))$ since this is a superset of $vertices(node(y))$. Otherwise, $node(y)$ lies in another part of the tree, and its label cannot contain $x$ (no node outside the subtree below $node(x)$ has $x$ in its label). Thus the hyperedge created when it is eliminated also cannot contain $x_i$.

In sum the hyperedge created when $x_i$ is eliminated is contained in $vertices(node(x_i))$, since all of the hyperedges containing $x_i$ at this stage are in this set. Furthermore, all vertices $x_1, \ldots, x_{i-1}$ are removed from this hyperedge, thus it contains only variables following $x_i$ in the ordering. Hence, by (1) above this hyperedge is contained in $label(node(x_i))$.  □

**Lemma 3** Let $\mathcal{H}$ be a hypergraph with elimination width at most $w$. Then $\mathcal{H}$ has a tree-decomposition of tree width at most $w$.

**Proof:** (Proof of Lemma 3) Let $\pi = x_1, \ldots x_n$ be an elimination ordering for $\mathcal{H}$. Then we will construct a tree decomposition for $\mathcal{H}$ using $\pi$ as follows. Initially, we have $|E|$ trees, each of size 1, one corresponding to each edge $e \in E$. We first merge the trees containing $x_n$ into a bigger tree, $T_n$, leaving us with a new, smaller set of trees. Then we merge the trees containing $x_{n-1}$ into a bigger tree, $T_{n-1}$. We continue in this way until we have formed a single tree, $T$. Now fill in the labels for all intermediate vertices of $T$ so that the tree is a tree-decomposition. That is, if $m$ and $n$ are two leaves of $T$ and they both contain some vertex $v$, then every node along the path from $m$ to $n$ must also contain $v$ in its label. It is not too hard to see that for each $x_i$, the tree $T_i$ (created when merging the trees containing $x_i$) has the property that the label of its root (which connects it with the rest of $T$) is contained in $e_i \cup x_i$, where $e_i$ is the hyperedge created when $x_i$ is eliminated. Basically, all nodes with $x_j, j > i$, are already contained in $T_i$ so $x_i$ does not need to label $T_i$'s root. Furthermore, if $x_j$ $j < i$ is contained in $T_i$'s root label, then $x_j$ must have been in some original hyperedge with a variable $x_k$ $k \geq i$: thus $x_j$ would have appeared in the hyperedge $e_i$ generated when $x_i$ was eliminated.

Hence the tree width of the final tree $T$ can be no larger than the induced width of $\pi$. $\square$

## A.2 Complexity Results for Caching Versions of DPLL

For the proof of theorems 1 and 2 we will need some common notation and definitions. Let $f$ be $k$-CNF formula with $n$ variables and $m$ clauses, let $\mathcal{H}$ be the underlying hypergraph associated with $f$ with branch width $w$. By the results of Darwiche (2001), there is a branch decomposition of $\mathcal{H}$ of depth $O(\log m)$ and width $O(w)$. Also by the results of Robertson and Seymour (1995), it is possible to find a branch decomposition, $T_{bd}$, such that $T_{bd}$ has branch width $O(w)$ and depth $O(\log m)$, in time $n^{O(1)} 2^{O(w)}$. Thus our main goal for each of the three theorems will be to prove the stated time and space bounds for our DPLL-based procedures, when they are run on a static ordering that is easily obtainable from $T_{bd}$.

Recall that the leaves of $T_{bd}$ are in one-to-one correspondence with the clauses of $f$. We will number the vertices of $T_{bd}$ according to a depth-first preorder traversal of $T_{bd}$. For a vertex numbered $i$, let $f_i$ denote the subformula of $f$ consisting of the conjunction of all clauses corresponding to the leaves of the tree rooted at $i$. Let $Vars(f_i)$ be the set of variables in the (sub)formula $f_i$. Recall that in a branch decomposition the label of each vertex $i$, $label(i)$, is the set of variables in the intersection of $Vars(f_i)$ and $Vars(f - f_i)$. Each node $i$ in $T_{bd}$ partitions the clauses of $f$ into three sets of clauses: $f_i$, $f_i^L$, and $f_i^R$, where $f_i^L$ is the conjunction of clauses at the leaves of $T_{bd}$ to the left of $f_i$, and $f_i^R$ is the conjunction of clauses at the leaves to the right of $f_i$.

All of our DPLL caching algorithms achieve the stated run time bounds by querying the variables in a specific, **static variable ordering**. That is, down any branch of the DPLL decision tree, $DT$, the same variables are instantiated in the same order. (In contrast a **dynamic variable ordering** allows DPLL to decide which variable to query next based on the assignments that have been made before. Thus different branches can query the variables in a different order.). The variable ordering used in $DT$ is determined by the depth-first pre-ordering of the vertices in the branch decomposition $T_{bd}$ and by the labeling of these vertices. Let $(i, 1), \ldots, (i, j_i)$ denote the variables in $label(i)$ that do not appear in the label of an earlier vertex of $T_{bd}$. Note that since the width of $T_{bd}$ is $w$, $j_i \leq w$ for all $i$. Let $1, \ldots, z$ be the sequence of vertex numbers of $T_{bd}$. Then our DPLL algorithm will query the variables underlying $f$ in the following static order: $\pi = \langle (i_1, 1), (i_1, 2), \ldots, (i_1, j_1), (i_2, 1), \ldots, (i_2, j_2), \ldots, (i_s, 1), \ldots, (i_s, j_s) \rangle$ $i_1 < i_2 < \ldots < i_s \leq z$, and $j_1, \ldots, j_s \leq w$.

Note that for some vertices $i$ of $T_{bd}$, nothing will be queried since all of the variables in its label may have occurred in the labels of earlier vertices. Our notation allows for these vertices to be skipped. The underlying complete decision tree, $DT$, created by our DPLL algorithms on input $f$ is thus a tree with $j_1 + j_2 + \ldots + j_s = n$ levels. The levels are grouped into $s$ layers, with the $i^{th}$ layer consisting of $j_i$ levels. Note that there are $2^l$ nodes at level $l$ in $DT$, and we will identify a particular node at level $l$ by $(l, \rho)$ where $\rho$ is a particular assignment to the first $l$ variables in the ordering, or by $((q, r), \rho)$, where $(q, r)$ is the $l^{th}$ pair in the ordering $\pi$, and $\rho$ is as before.

The DPLL algorithms carry out a depth-first traversal of $DT$, keeping formulas in the cache that have already been solved along the way. (For #DPLL-SimpleCache, the formulas stored in the cache are of the form $f|_\rho$, and for #DPLL-Cache and #DPLL-Space, the formulas stored are various components of ToComponents($f|_\rho$).) If the algorithm ever hits a node where the formula to be computed has already been solved, it can avoid that computation, and thus it does not do a complete depth-first search of $DT$ but rather it does a depth-first search of a *pruned* version of $DT$. For our theorems, we want to get an upper bound on the size of the pruned tree actually searched by the algorithm.

**Theorem 1** For solving #SAT with $n$ variables, there is an execution of #DPLL-SimpleCache that runs in time bounded by $2^{O(w \log n)}$ where $w$ is the underlying branch width of the instance. Furthermore, the algorithm can be made deterministic with the same time guarantees.

**Proof:** We want to show that the size of the subtree of $DT$ searched by #DPLL-SimpleCache is at most $2^{O(w \log n)}$. When backtracking from a particular node $(l, \rho) = ((q, r), \rho)$ at level $l$ in $DT$, the formula put in the cache, if it is not already known, is of the form $f|_\rho$. (Recall $\rho$ is a setting to the first $l$ variables.) However, we will see that although there are $2^l$ different ways to set $\rho$, the number of *distinct* formulas of this form is actually much smaller than $2^l$. Consider a partial assignment, $\rho$, where we have set all variables up to and including $(q, r)$, for some $q \leq i_s$ and some $r \leq j_q$. The number of variables set by $\rho$ (the *length* of $\rho$) is $j_1 + j_2 + \ldots + j_{q-1} + r$.

Let $\rho^-$ denote the partial assignment that is consistent with $\rho$ where only the variables in $\rho$ that came from the labels of the vertices on the path from the root of $T_{bd}$ up to and including vertex $q$ are set. The idea is that $\rho^-$ is a reduction of $\rho$, where $\rho^-$ has removed the assignments of $\rho$ that are irrelevant to $f_q$ and $f_q^R$.

Consider what happens when the DPLL algorithm reaches a particular node $((q, r), \rho)$ at level $l$ of $DT$. At that point the algorithm is solving the subproblem $f|_\rho$, and thus, once we backtrack to this node, $f|_\rho = f_q^L|_\rho \wedge f_q|_\rho \wedge f_q^R|_\rho$ is placed in the cache, if it is not already known. Note that all variables in the subformula $f_q^L$ are set by $\rho$, and thus either $f_q^L|_\rho = 0$, in which case nothing new is put in the cache, or $f_q^L|_\rho = 1$ in which case $f|_\rho = f_q|_\rho \wedge f_q^R|_\rho = f_q|_{\rho^-} \wedge f_q^R|_{\rho^-}$ is put in the cache. Thus, the set of *distinct* subformulas placed in the cache at level $l = (q, r)$ is at most the set of all subformulas of the form $f_q|_{\rho^-} \wedge f_q^R|_{\rho^-}$, where $\rho^-$ is a setting to all variables in the labels from the root to vertex $q$, plus the variables $(q, 1), \ldots, (q, r)$. There are at most $d \cdot w$ such variables, where $q$ has depth $d$ in $T_{bd}$ (each label has at most $w$ variables since this is the width of $T_{bd}$). Hence the total number of such $\rho^-$'s is at most $2^{(w \cdot d)}$. This implies that the number of subtrees in $DT$ at level $l + 1$ that are actually traversed by #DPLL-SimpleCache is at most $2 \cdot 2^{w \cdot d} = 2^{O(w \cdot d)}$, where $d$ is the depth of node $q$ in $T_{bd}$. Let $t$ be the number of nodes in $DT$ that are actually traversed by #DPLL-SimpleCache. Then, $t$ is at most $n 2^{O(w \cdot \log n)}$, since $t$ is the sum of the number of nodes visited at every level of $DT$ and for each node $q$ in $T_{bd}$ $d \in O(\log m) = O(\log n)$.

Accounting for the time to search the cache, the overall runtime of #DPLL-SimpleCache is at most $t^2$, where again $t$ is the number of nodes in $DT$ that are traversed by the algorithm. Thus, #DPLL-SimpleCache runs in time $(n2^{O(w \cdot \log n)})^2 = 2^{O(w \cdot \log n)}$. □

**Theorem 2** *For solving #SAT on $n$ variables, there exists an execution of #DPLL-Cache that runs in time bounded by $n^{O(1)}2^{O(w)}$ where $w$ is the underlying branch width of the instance. Furthermore, the algorithm can be made deterministic with the same time guarantees.*

**Proof:** We prove the theorem by placing a bound on the number of times #DPLL-Cache can branch on any variable $x_l$. Using the notation specified above, $x_l$ corresponds to some pair $(q, r)$ in the ordering $\pi$ used by #DPLL-Cache. That is, $x_l$ is the $r$'th new variable in the label of vertex $q$ of the branch decomposition $T_{bd}$.

When #DPLL-Cache utilizes the static ordering $\pi$, it branches on, or queries, the variables according to that order, always reducing the component containing the variable $x_i$ that is currently due to be queried. However, since previously cached components are always removed (by RemoveCachedComponents in the algorithm), it can be that when it is variable $x_i$'s turn to be queried, there is no component among the active components that contains $x_i$. In this case, #DPLL-Cache simply moves on to the next variable in the ordering, continuing to advance until it finds the first variable that does appear in some active component. It will then branch on that variable reducing the component it appears in, leaving the other components unaltered.

This implies that at any time when #DPLL-Cache selects $x_l$ as the variable to next branch on it must be the case that (1) $x_l$ appears in an active component. In particular the value of this component is not already in the cache. And (2) no variable prior to $x_l$ in the ordering $\pi$ appears in an active component. All of these variables have either been assigned a particular value by previous recursive invocations, or the component they appeared in has been removed because its value was already in the cache.

In the branch decomposition $T_{bd}$ let $p$ be $q$'s parent ($q$ must have a parent since the root has an empty label). We claim that whenever #DPLL-Cache selects $x_l$ as the next variable to branch on, the active component containing $x_l$ must be a component in the reduction of $f_p$ whose form is determined solely by the settings of the variables in $p$ and the $r$ variables of $q$ that have already been set. If this is the case, then there can be at most $2^{(w+r)} = 2^{O(w)}$ different components that $x_l$ can appear in, and hence #DPLL-Cache can branch on $x_l$ at most $2^{O(w)}$ times as each time one more of these components gets stored in the cache.

Now we prove the claim. The label of $q$ consists of variables appearing in $p$'s label and variables appearing in the label of $q$'s sibling. Since all of the variables in $label(p)$ have been set, $q$ and its sibling must now have an identical set of unqueried variables in their labels. Hence, $q$ must be the left child of $p$ as by the time the right child is visited in the ordering, $x_l$ will have already been queried. Thus, at the time $x_l$ is queried, $f_p$ will have been affected only by the current setting of $label(p)$ (as these are the only variables it shares with the rest of the formula) and the first $r$ queried variables from $label(q)$. That is, $f_p$ can be in at most $2^{(w+r)}$ different configurations, and thus the component containing $x_l$ can also be in at most this many different configurations.

Thus with $n$ variables we obtain a bound on the number of branches in the decision tree explored by #DPLL-Cache of $n2^{O(w)}$. As in the proof of the previous theorem, the overall runtime is at most quadratic in the number of branches traversed, to give the claimed bound of $n^{O(1)}2^{O(w)}$. □

**Theorem 3** For solving #SAT on $n$ variables, there is an execution of #DPLL-Space that uses only space linear in the instance size and runs in time bounded by $2^{O(w \log n)}$ where $w$ is the underlying branch width of the instance. Furthermore, the algorithm can be made deterministic with the same time and space guarantees.

**Proof:** For this proof, it will be more natural to work with a *tree decomposition* rather than a branch decomposition.

Let $f$ be a $k$-CNF formula with $n$ variables and $m$ clauses and let $\mathcal{H}$ be the underlying hypergraph associated with $f$. We begin with a tree decomposition $T_{td}$ of depth $O(\log m)$ and width $O(w)$ (computable in time $n^{O(1)}2^{O(w)}$). We can assume without loss of generality that the leaves of $T_{td}$ are in one-to-one correspondence with the clauses of $f$. Each node $i$ in $T_{td}$ partitions $f$ into three disjoint sets of clauses: $f_i$, the conjunction of clauses at the leaves of the subtree of $T_{td}$ rooted at $i$, $f_i^L$, the conjunction of clauses of the leaves of $T_{td}$ to the left of $f_i$, and $f_i^R$, the conjunction of clauses of the leaves of $T_{td}$ to the right of $f_i$. #DPLL-Space will query the variables associated with the labels of $T_{td}$ according to the depth-first preorder traversal. Let the variables in $label(i)$ not appearing in an earlier label on the path from the root to node $i$ be denoted by $S(i) = (i, 1), \ldots, (i, j_i)$. If $i$ is a non-leaf node with $j$ and $k$ being its left and right children, then the variables in $S(i)$ are exactly the variables that occur in both $f_j$ and $f_k$ but that do not occur outside of $f_i$. If we let $c$ be the total number of nodes in $T_{td}$, then #DPLL-Space will query the variables underlying $f$ in the following static order: $S(1), S(2), \ldots, S(c)$, where some $S(i)$ may be empty. The underlying decision tree, $DT$, created by #DPLL-Space is a complete tree with $n$ levels. As before we will identify a particular node $s$ at level $l$ of $DT$ by $s = (l, \rho)$ where $\rho$ is a particular assignment to the first $l$ variables in the ordering, or by $s = ((q, r), \rho)$ (the $r^{th}$ variable in $S(q)$).

#DPLL-Space carries out a depth-first traversal of $DT$, storing the components of formulas in the cache as they are solved. However, now components of formulas are also popped from the cache so that the total space ever utilized is linear. If the algorithm hits a node where all of the components of the formula to be computed are known, it can avoid traversing the subtree rooted at that node. Thus it searches a pruned version of $DT$.

During the (pruned) depth-first traversal of $DT$, each edge that is traversed is traversed twice, once in each direction. At a given time $t$ in the traversal, let $E = E_1 \cup E_2$ be the set of edges that have been traversed, where $E_1$ are the edges that have only been traversed in the forward direction, and $E_2$ are the edges that have been traversed in both directions. The edges in $E_1$ constitute a partial path $p$ starting at the root of $DT$. Each edge in $p$ is labeled by either 0 or 1. Let $p_1, \ldots, p_k$ be the set of all subpaths of $p$ (beginning at the root) that end in a 1-edge. Let $\rho_1, \ldots, \rho_k$ be substrictions corresponding to $p_1, \ldots, p_k$ except that the last variable that was originally assigned a 1 is now assigned a 0. For example, if $p$ is $(x_1 = 0, x_3 = 1, x_4 = 0, x_5 = 1, x_6 = 0, x_2 = 0)$, then $\rho_1 = (x_1 = 0, x_3 = 0)$, and $\rho_2 = (x_1 = 0, x_3 = 1, x_4 = 0, x_5 = 0)$. Then the information that is in the cache at time $t$ contains ToComponents($f|_{\rho_i}$), $i \leq k$.

For a node $q$ of $T_{td}$ and corresponding subformula $f_q$, the *context* of $f_q$ is a set of variables defined as follows. Let $(q_1, \ldots, q_d)$ denote the vertices in $T_{td}$ on the path from the root to $q$ (excluding $q$ itself). Then the context of $f_q$ is the set $Context(f_q) = S(q_1) \cup S(q_2) \cup \ldots \cup S(q_d)$. Intuitively, the context of $f_q$ is the set of all variables that are queried at nodes that lie along the path to $q$. Note that when we reach level $l = (q, 1)$ in $DT$, where the first variable of $S(q)$ is queried, we have already queried many variables, including all the variables in $Context(f_q)$. Thus the set of all variables queried up to level $l = (q, 1)$ can be partitioned into two groups relative to $f_q$: the irrelevant variables, and the set $Context(f_q)$ of relevant variables. We claim that at an arbitrary level $l = (q, r)$ in $DT$, the only nodes at level $l$ that are actually traversed are those nodes $((q, r), \rho)$ where all irrelevant variables in $\rho$ (with respect to $f_q$) are set to 0. The total number of such nodes at level $l = (q, r)$ is at most $2^{|Context(f_q)|+r}$ which is at most $2^{w \log n}$. Since this will be true for all levels, the total number of nodes in $DT$ that are traversed is bounded by $n2^{w \log n}$. Thus, all that remains is to prove our claim.

Consider some node $s = ((q, r), \alpha)$ in $DT$. That is, $\alpha = \alpha^1 \alpha^2 \ldots \alpha^{q-1} b_1 \ldots b_{r-1}$, where for each $i$, $\alpha^i$ is an assignment to the variables in $S(i)$, and $b_1 \ldots b_{r-1}$ is an assignment to the first $r - 1$ variables in $S(q)$. Let the context of $f_q$ be $S(q_1) \cup \ldots \cup S(q_d)$, $d \leq \log n$. Now suppose that $\alpha$ assigns a 1 to some non-context (irrelevant) variable, and say the first such assignment occurs at $\alpha_t^u$, the $t^{th}$ variable in $\alpha^u$, $u \leq q - 1$. We want to show that the algorithm never traverses $s$.

Associated with $\alpha$ is a partial path in $DT$; we will also call this partial path $\alpha$. Consider the subpath/subassignment $p$ of $\alpha$ up to and including $\alpha_t^u = 1$. If $\alpha$ is traversed, then we start by traversing $p$. Since the last bit of $p$ is 1 (i.e., $\alpha_t^u = 1$) when we get to this point, we have stored in the cache ToComponents$(f|_\rho)$ where $\rho$ is exactly like $p$ except that the last bit, $\alpha_t^u$, is zero. Let $j$ be the first node in $q_1, q_2, \ldots q_d$ with the property that the set of variables $S(j)$ are not queried in $p$. (On the path to $q$ in $T_{td}$, $j$ is the first node along this path such that the variables in $S(j)$ are not queried in $p$.) Then ToComponents$(f|_\rho)$ consists of three parts: (a) ToComponents$(f_j^L|_\rho)$, (b) ToComponents$(f_j|_\rho)$, and (c) ToComponents$(f_j^R|_\rho)$.

Now consider the path $p'$ that extends $p$ on the way to $s$ in $DT$, where $p'$ is the shortest subpath of $\alpha$ where all of the variables $S(i)$ for $i < j$ have been queried. The restriction corresponding to $p'$ is a refinement of $p$ where all variables in $S(1) \cup S(2) \cup \ldots S(j-1)$ are set. Since we have already set everything that occurs before $j$, we will only go beyond $p'$ if some component of ToComponents$(f|_{p'})$ is not already in the cache. ToComponents$(f|_{p'})$ consists of three parts: (a) ToComponents$(f_j^L|_{p'})$, (b) ToComponents$(f_j|_{p'})$, and (c) ToComponents$(f_j^R|_{p'})$. Because we have set everything that occurs before $j$, all formulas in (a) will be known. Since $p'$ and $\rho$ agree on all variables that are relevant to $f_j$, ToComponents$(f_j|_{p'}) = $ ToComponents$(f_j|_\rho)$ and hence these formulas in (b) in the cache. Similarly all formulas in (c) are in the cache since ToComponents$(f_j^R|_{p'}) = $ ToComponents$(f_j^R|_\rho)$. Thus all components of ToComponents$(f|_{p'})$ are in the cache, and hence we have shown that we never traverse beyond $p'$ and hence never traverse $s$. Therefore the total number of nodes traversed at any level $l = (q, r)$ is at most $2^{wd}$, where $d$ is the depth of $q$ in $T_{td}$, as desired. This yields an overall runtime of $2^{O(w \log n)}$.

It is left to argue that the space used is linear in the instance size. The total number of formulas that are ever stored in the cache simultaneously is linear in the depth of the tree decomposition, which is $O(\log m)$. Since we store each restricted formula $f|_\rho$ by storing the associated restriction $\rho$, the total space ever used is $O(n \log m)$, which is linear in the input size. □

### A.3 Comparing Algorithms for BAYES and #SAT

Before proving the next theorem, we first discuss in more detail the structure of the search space explored by various versions of RC, AND/OR search and DDP. All of these algorithms operate in the same way. They instantiate variables and when the problem decomposes into independent components they solve these components in separate recursions. Hence, when solving any CNF formula $f$ they all generate some AND/OR search tree (Dechter & Mateescu, 2007).

The AND/OR search tree $AO$ generated when one of the above algorithms solves the #SAT instance $f$ (a CNF formula), is a rooted tree. Each node $n$ of $AO$ is labeled by a formula $n.f$ and the subtree below $n$ is generated when solving $n.f$. The root of $A0$ is labeled by the original formula $f$. There are four different types of nodes in $AO$:

**Query nodes.** Each query node $q$ has an associated variable $q.var$ and two children corresponding to the two possible instantiations of $q.var$. That is, its children are labeled by the formulas $q.f|_{q.var=0}$ and $q.f|_{q.var=1}$. A query node $q$ is generated by the search algorithm whenever it chooses to instantiate $q.var$ and then executes recursive calls on the two resultant reduced formulas.

**AND nodes.** Each AND node, $a$, has a query node as its parent, and has one or more children all of which are query nodes. An AND node is generated by the search algorithm when it decomposes the current formula into two or more independent components following the instantiation of the parent query node's variable. Each of these components will then be solved in one of the subtrees rooted by the AND node's children. If $a.f$ splits into the components $f_i$, $i = 1, \ldots, k$, then $a.f = \bigwedge_i f_i$, and the $i$'th child of $a$ is labeled by $f_i$. Note that the $f_i$ share no variables. Hence, the set of query node variables that appear in the subtree below the $i$-th child of $a$ are disjoint from the set of query node variables appearing below the $j$-th child of $a$ for all $j \neq i$.

**Failure nodes.** These are leaf nodes of the tree that are labeled with a formula containing the empty clause. If caching is being used, failure nodes might also be labeled by a formula in the cache that has already been shown to be unsatisfiable.

**Satisfying nodes.** These are leaf nodes of the tree that are labeled with a formula containing no clauses. If caching is being used, satisfying nodes might also be labeled by a satisfiable formula in the cache whose model count is already know.

Figure 6 shows an example AND/OR search tree.

Each node $n$ of the $AO$ also has a value, $n.value$, computed by the algorithm that generates it. Here we only need to distinguish between zero values $n.value = 0$, and non-zero values denoted by $n.value = 1$. Every satisfying node has value 1, and every failure node has value 0. A query node has value 1 if and only if at least one of its children has value 1, and an AND node has value 1 if and only if all of its children have value 1. For example, in Figure 6 AND node D has value 0, while query node 2 has value 1. Note that all of the children of an AND node in $AO$ must have value 1 except possibly the right most child. The algorithms generating $AO$ all terminate the search below an AND node as soon as they discover a value 0 child—this implies that the AND node has value 0. It can be seen that $n.value = 0$ if $n.f$ is unsatisfiable and $n.value = 1$ if $n.f$ is satisfiable.

Given any node $n$ of $AO$, let $AO(n)$ be the AND/OR subtree of $AO$ rooted by $n$. Each satisfying assignment $\rho$ of $n$'s formula $n.f$ defines a **solution subtree** $S(n)$ of $AO(n)$. In particular, $S(n)$ is a connected subtree of $AO(n)$ rooted by $n$ such that (1) if $q$ is a query node in $S(n)$ then $S(n)$ also contains the child of $q$ corresponding to the assignment made by $\rho$ (i.e., if $\rho[q.var]$ is the value assigned to $q.var$ in $\rho$, then $S(n)$ will contain the child labeled by the formula $q.f|_{q.var=\rho[q.var]}$), (2) if $a$ is an AND node in $S(n)$ then $S(n)$ contains all children of $a$, and (3) $S$ contains no failure nodes. For example, a solution subtree of the AND/OR tree shown in Figure 6 (i.e., a solution subtree of the root node) is formed by the leaf nodes **b**, **c**, **f**, and **l**; the query nodes 1, 2, 3, 4, 5, 6, 7, and 8; and the AND nodes A, and B. In particular, the left value of query nodes 1, 2, 3, 5, 6, 7 and 8, along with the right value of query node 4 satisfy all clauses of the formula $1.f$. A solution subtree of $AO(n)$ exists if and only if $n.value = 1$.

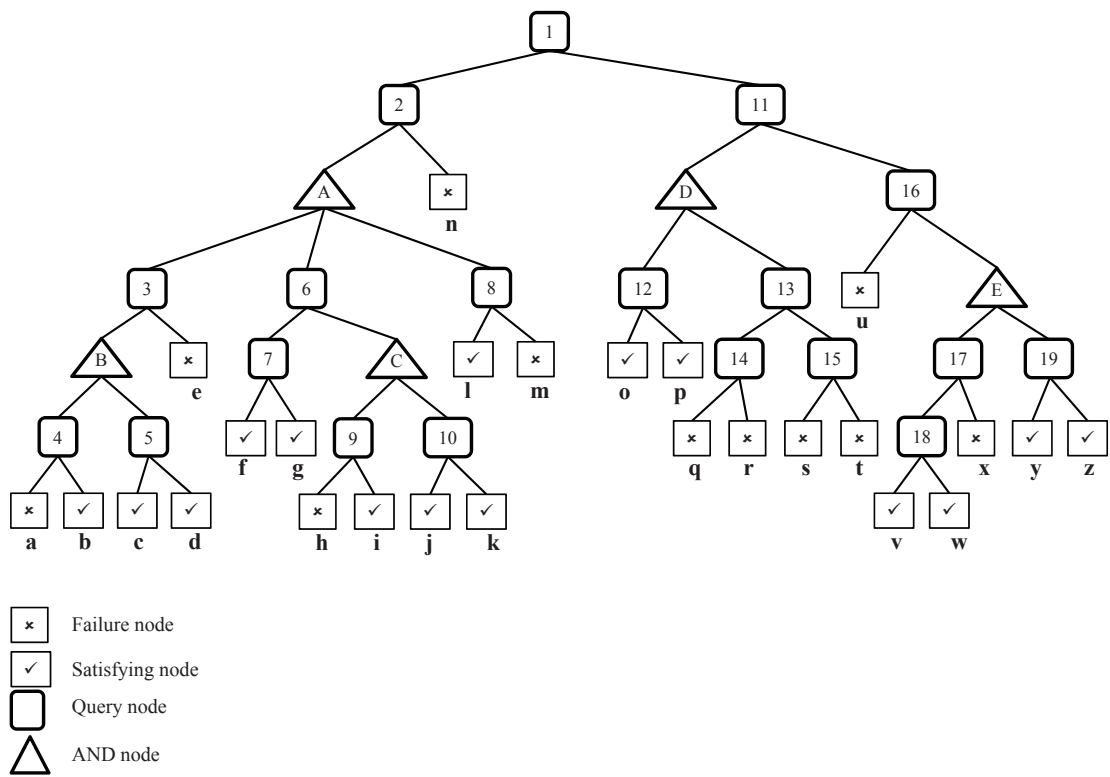

Figure 6: An example AND/OR search tree with query nodes numbered 1–19, leaf nodes (both failure and satisfying) numbered a–z, and AND nodes labeled A–E.

Finally, in an AND/OR search tree we say that a query node whose parent is an AND node is a **component root**. We also classify the root node as a component root. In Figure 6 query nodes 1 (the root node), 3, 6, 8, 4, 5, 9, 10, 12, 13, 17, and 19 are component roots.

**Theorem 5** #DPLL-Cache polynomially simulates RC-Cache, RC-Cache$^+$, AND/OR-Cache, AND/OR-Cache$^+$, and VE. #DPLL-Space polynomially simulates RC-Space, AND/OR-Space and DDP.

**Proof:** Since RC-Cache polynomially simulates VE we can ignore VE in our proof: showing that #DPLL-Cache polynomially simulates RC-Cache also shows that it polynomially simulates VE. Also we assume in our proof that if any of these algorithms use unit propagation, then so does #DPLL-Cache/Space. As explained in Section 3.1, #DPLL-Cache/Space without unit propagation can polynomially simulate versions of #DPLL-Cache/Space using unit propagation.

Each of the stated algorithms will generate an AND/OR search tree when solving a CNF formula $f$. To prove the theorem we first show how any AND/OR search tree solving $f$ can be converted into a partial DPLL decision tree, $DT$, that is no bigger. Then we show that our DPLL algorithms can solve $f$ using $DT$ to guide its variable ordering. Thus, we obtain the result that the minimal runtime for any of the stated algorithms, which must result in the generation of some AND/OR search tree $AO_{\min}$, can also be achieved by our DPLL algorithms. In particular, when run on the partial decision tree constructed from $AO_{\min}$, our DPLL algorithms will achieve a polynomially similar runtime. (This suffices to prove the theorem, as we need only show the **existence** of an execution of our DPLL algorithms achieving this run time.)

To make the distinction between the AND/OR search tree and the constructed partial decision tree clear, we will use the suffixes $_{ao}$ and $_{dt}$ to indicate elements of the AND/OR tree and decision tree respectively.

DPLL decision trees contain only query variables, satisfying nodes, and failure nodes, where satisfying and failure nodes are both leaf nodes. We construct a partial decision tree $DT$ from an AND/OR tree $AO$ by expanding the left most solution subtree $S(n_{ao})$ below every node $n_{ao} \in AO$ with $n_{ao}.value = 1$ into a linear sequence of query variables in $DT$ using a depth-first ordering of the query variables in $S(n_{ao})$. For nodes $n_{ao} \in AO$ with $n_{ao}.value = 0$ the same expansion is attempted, but in this case it will result in a sequence of query nodes that terminate at failure nodes.

Every node $q_{dt}$ in $DT$ has a pointer, $dt{\to}ao(q_{dt})$ to a node $q_{ao}$ in $AO$, at the end of the construction these pointers establish a map between the nodes in $DT$ and the nodes in $AO$. Initially, the root of $DT$ has a pointer to the root of $AO$. Then, for any node $q_{dt} \in DT$:

1. If $dt{\to}ao(q_{dt})$ is a query node $q_{ao}$ in $AO$, then make $q_{dt}$ a query node and create a left and right child, $l_{dt}$ and $r_{dt}$, for $q_{dt}$ in $DT$. We make $q_{dt}$ query the same variable as $q_{ao}$ (i.e., $q_{dt}.var = q_{ao}.var$), and set its children to point to the children of $q_{ao}$ (i.e., $dt{\to}ao(l_{dt})$ and $dt{\to}ao(r_{dt})$ are set to the left and right children of $q_{ao}$ in $AO$).

2. If $dt{\to}ao(q_{dt})$ is an AND node $a_{ao}$ in $AO$, then we reset $dt{\to}ao(q_{dt})$ to be the left most child of $a_{ao}$ in $AO$. We then apply the first rule above, and continue.

3. If $dt{\to}ao(q_{dt})$ is a failure node in $AO$ then we set $q_{dt}$ to be a failure node. In this case $q_{dt}$ has no children.

4. If $dt{\to}ao(q_{dt})$ is a satisfying node in $AO$ then we examine the path $\rho_{ao}$ in $AO$ from the root to $dt{\to}ao(q_{dt})$. Let $r_{ao}$ be the last component root on $\rho_{ao}$ that has a right sibling.

   (a) If such an $r_{ao}$ exists, and no node on the path from $r_{ao}$ to $dt{\to}ao(q_{dt})$ in $AO$ is the right child of a query node whose left child has value 1, then we reset $dt{\to}ao(q_{dt})$ to be the leftmost right sibling of $r_{ao}$. This node is also a component root, and hence it is a query node in $AO$. We then apply the first rule above, and continue.

   (b) Otherwise (either $r_{ao}$ does not exist or there is some node on the path from $r_{ao}$ that is the right child of a query node whose left child has value 1), we make $q_{dt}$ a satisfying node. In this case $q_{dt}$ has no children.

Rule 4 of the construction is where we convert the leftmost solution subtree below each node $n_{ao}$ in $AO$ into a sequence of query nodes in $DT$ by performing a depth-first traversal of this solution subtree. In particular, in this solution subtree the leftmost right sibling of the deepest component root is the depth-first successor of the satisfying leaf node. The condition that no node on route to that sibling is the right child of a query node whose left child has value 1 ensures that we only perform a depth-first traversal along the leftmost solution subtree and not along subsequent solution subtrees. Figure 7 shows the partial decision tree that would be constructed from the AND/OR search tree of Figure 6.

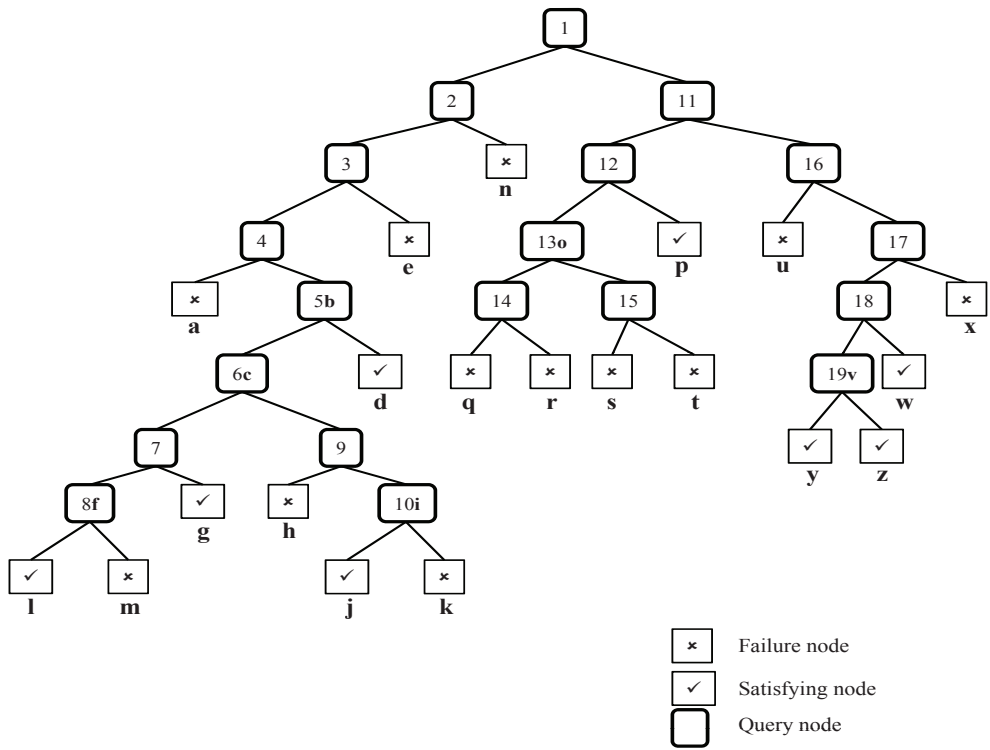

Figure 7: The partial DPLL decision tree constructed from the AND/OR search tree of Figure 6. Each query and leaf nodes $n$ is numbered with the number of the corresponding node, $dt{\rightarrow}ao(n)$, in the AND/OR search tree.

In the diagram, satisfying nodes whose pointers are reset to the next component root using rule 4a, are numbered with the corresponding query node in the AND/OR tree followed by the leaf label of the corresponding satisfying node. For example, node **5b** in the Figure 7 represents satisfying child **b** of node 4 in Figure 6 that has been redirected to its depth-first successor node 5 (the leftmost right sibling of the deepest component root 4).

As another example, in the AND/OR tree, the right child of node 6 is the AND node C. Hence, in the decision tree, the right child of the corresponding query node 6, becomes query node 9 which is the leftmost child of node C (rule 2). Furthermore, when we reach satisfying node **j** in the AND/OR tree, we can proceed no further and hence the left child of query node 10 in the decision tree becomes a terminal satisfying node (rule 3). In particular, although the path from the root to node 10 in the AND/OR tree contains a component root with a right sibling, namely node 6, this path also contains the node C that is the right child of a query node (node 6) whose left child (node 7) has value 1.

There are two things to note. First, at any node $n_{dt}$ of $DT$ all variables instantiated on the path $\rho_{ao}$ in $AO$ from the root to $dt{\rightarrow}ao(n)$ have been instantiated to the same values on the path $\rho_{dt}$ in $DT$ from the root to $n_{dt}$. Since Rules 3 and 4b terminate paths, all nodes on $\rho_{dt}$ are inserted only by Rules 1, 2, and 4b. Rules 1 and 2 only insert nodes on $\rho_{dt}$ whose parents are already on $\rho_{dt}$, and Rule 1 ensures that the values assigned are the same as those in $AO$. Finally, Rule 4a only inserts a node $a_{dt}$ on $\rho_{dt}$ if one of $dt{\rightarrow}ao(a_{dt})$'s siblings is already on $\rho_{dt}$, and hence that sibling's (and $a$'s) parent must already be on $\rho_{dt}$.

Second, no variable is queried twice along any path of $DT$. That is, no node $n_{dt}$ in $DT$ has an ancestor $n'_{dt}$ with $n_{dt}.var = n'_{dt}.var$. Again any path $\rho_{dt}$ in $DT$ is grown only by applications of Rules 1, 2, and 4a. Since no path in $AO$ queries the same variable twice, Rules 1 and 2 must preserve this condition. Similarly Rule 4a moves to a new component root $a_{ao}$, and the set of query variables at and below $a_{ao}$ in $AO$ is disjoint with the set of query variables already appearing in $\rho_{dt}$.

Using the above, from the AND/OR search tree $AO$ generated by any of the algorithms RC-Space, AND/OR-Space or DDP when solving the formula $f$, we can construct a corresponding partial decision tree $DT$. Now we show that #DPLL-Space can solve $f$ by exploring a search tree that is no larger than $DT$. Note that $DT$ is itself no larger than $AO$, hence this will show that #DPLL-Space can solve $f$ with a polynomially similar run time, proving that it can polynomially simulate RC-Space, AND/OR-Space and DDP. (Note that the run time of all of these algorithms is polynomially related to the size of the search trees they explore.)

We execute #DPLL-Space using the variable ordering specified in $DT$. That is, starting at the root $r_{dt}$ of $DT$, #DPLL-Space will always query the variable of the current node of $DT$, $n_{dt}.var$, and then descend to $n_{dt}$'s left child. When it backtracks to $n_{dt}$ it will then descend to the right child. Hence, we only need to show that #DPLL-Space must backtrack if it reaches a leaf of $DT$. That is, it explores a search tree that is no larger than $DT$.

First, if #DPLL-Space reaches a failure node of $DT$ it must detect an empty clause and backtrack. By Rule 3 of the construction any failure node $f_{dt}$ of $DT$ must correspond to a failure node $dt{\rightarrow}ao(f_{dt})$ in $AO$. Since all variables instantiated on the path in $AO$ from the root to $dt{\rightarrow}ao(f_{dt})$ are instantiated to the same values on the path in $DT$ from the root to $f_{dt}$, we see that if an empty clause was detected in $AO$ at $dt{\rightarrow}ao(f_{dt})$ then #DPLL-Space must also detect an empty clause at $f_{dt}$. (Note that if the algorithm that generated $AO$ used unit propagation, then we assume that #DPLL-Space does as well).

Second, if #DPLL-Space reaches a satisfying node $s_{dt}$ of $DT$ it must detect that all of its current set of components are solved and backtrack (line 4 of Algorithm 8). Let $\rho_{dt}$ be the path in $DT$ from the root to $s_{dt}$, $\rho_{ao}$ be the path in $AO$ from $dt{\rightarrow}ao(s_{dt})$ to the root, and $cr_{dt}$ be a node on $\rho_{dt}$ such that $dt{\rightarrow}ao(cr_{dt})$ is a component root in $AO$ (we say that $cr_{dt}$ is a component root on $\rho_{dt}$). We claim that (a) if $l_{ao}$ is a left sibling of $dt{\rightarrow}ao(cr_{dt})$ in $AO$, then there exists a node $l_{dt}$ on $\rho_{dt}$ such that $dt{\rightarrow}ao(l_{dt}) = l_{ao}$, and $l_{ao}.f$ is satisfied by $\rho_{dt}$; (b) if $r_{ao}$ is a right sibling of $dt{\rightarrow}ao(cr_{dt})$ in $AO$ then $r_{ao}.f$ is in #DPLL-Space's cache.

Given claim (a) the only clauses of the original formula not yet satisfied by $\rho_{dt}$ are clauses from $r_{ao}.f$ for those nodes $r_{ao}$ in $AO$ that are right siblings of some component root $cr_{dt}$ on $\rho_{dt}$ (i.e., $r_{ao}$ is a right sibling of component root $dt{\rightarrow}ao(cr_{dt})$ in $AO$). When #DPLL-Space arrived at $cr_{dt}$, prior to reaching $s_{dt}$, all variables in $AO$ on the path from the root to $dt{\rightarrow}ao(cr_{dt})$ have already been instantiated to the same values on $\rho_{dt}$. Thus, if $p_{ao}$ is $dt{\rightarrow}ao(cr_{dt})$'s parent in $AO$, #DPLL-Space would have recognized that $r_{ao}.f$ was a separate component once it instantiated $p_{ao}.var$, and it would have added $r_{ao}.f$ to its list of components (at line 8 or 11 of Algorithm 8). Note that, once solved $r_{ao}.f$ would not be removed from #DPLL-Space's cache until it backtracks to undo the instantiation of $p_{ao}.var$. (At which point the solution all of $p_{ao}$'s children would be combined to yield a solution to $p_{ao}.f$).

Furthermore, following the variable ordering specified in $DT$, #DPLL-Space would not instantiate any of the variables in $r.f$ along the path $\rho_{dt}$. Hence, any component that is on #DPLL-Space's list of components when it reaches $n_s$ must be equal to $r_{ao}.f$ for some right sibling $r_{ao}$ of a component root on $\rho_{dt}$, and by claim (b) will be removed by the call to RemoveCachedComponents($\Phi$) (line 6). This will leave #DPLL-Space with an empty list of components to solve, and hence it must backtrack at $s_{dt}$.

Now we prove the claims. For (a) we see that $DT$ will always visit the children of an AND node in $AO$ in a left to right order. That is, before inserting a component root $cr_{dt}$ on its path, it must first visit all left siblings $l_{ao}$ of $dt{\rightarrow}ao(cr_{dt})$. After inserting $l_{dt}$ on its path (with $dt{\rightarrow}ao(l_{dt}) = l_{ao}$), it will instantiate $l_{dt}$ and then start to query the nodes under $l_{ao}$ searching alternate instantiations to these variables until it is able to traverse a leftmost solution subtree of $AO(l_{ao})$. This traversal results in the insertion into the path of a solution to $l_{ao}.f$, after which $DT$ inserts $cr_{dt}$ on its path using Rule 4a.

For (b) we observe that $s_{dt}$ is a satisfying node in $DT$ only through the application of Rule 4b. Hence there are two possible cases. First, it can be that none of the component roots on $\rho_{dt}$ have a right sibling. In this case every clause of the original formula is satisfied and #DPLL-Space must backtrack. For example, in Figure 7 this occurs at leaf nodes **l** and **y**.

Otherwise, let $cr_{dt}$ be a component root on $\rho_{dt}$ such that $dt{\rightarrow}ao(cr_{dt})$ has a right sibling in $AO$, and let $n_{dt}$ be the first node on $\rho_{dt}$ following $cr_{dt}$ such that (i) $n_{dt}$'s successor on $\rho_{dt}$ is its right child, and (ii) $dt{\rightarrow}ao(n_{dt})$ has a left child in $AO$ with value 1. Such a node $n_{dt}$ must exist, else $s_{dt}$ would not have been a leaf node of $DT$ by Rule 4a. When #DPLL-Space arrived at node $cr_{dt}$ it would have $r_{ao}.f$ on its list of components for all right siblings $r_{ao}$ of $dt{\rightarrow}ao(cr_{dt})$. There might also be other unsolved components on this list. All of these components, however, must be equal to $r_{ao}.f$ for some right sibling $r_{ao}$ of a component root on $\rho_{dt}$ preceding $cr_{dt}$, and must have been placed on the list of components prior to #DPLL-Space reaching $cr_{dt}$. Then, when #DPLL-Space arrived at $n_{dt}$ it would have taken the left branch first. Thus it would have previously been invoked with all of these right sibling components on its component list.

When #DPLL-Space is invoked with a list of components it either solves every component, placing them in its cache and keeping them there until it backtracks to the node where they were first placed on its list, or it discovers that one of these components is unsatisfiable. If one of the components is unsatisfiable, it will immediately backtrack to the point where that component was first placed on its list. In particular, all recursive calls where the list of components contains a known unsatisfiable component will return immediately since the call to InCache($\Phi$) will detect that the list of components has product equal to zero.

Hence, on taking the left branch at $n_{dt}$, #DPLL-Space, will have on its list of components, components of the form $r_{ao}.f$ for right siblings of component roots above $n_{dt}$ on $\rho_{dt}$, and also $l_{ao}.f$ where $l_{ao}$ is the left child of $dt{\rightarrow}ao(n_{dt})$ in $AO$. Since $l_{ao}$ has value 1, $l_{ao}.f$ is satisfiable, and either #DPLL-Space will solve all its components, placing their value in its cache, or it will discover that one of the components $r_{ao}.f$ is unsatisfiable and will backtrack without visiting $s_{dt}$. Therefore, if it does visit $s_{dt}$ it would have solved all components that could potentially be on its list of components, and these components would still be in is cache since they were placed on the list before arriving at $s_{dt}$.

This shows that #DPLL-Space polynomially simulates RC-Space, AND/OR-Space and DDP. RC-Cache and AND/OR-Cache gain over RC-Space and AND/OR-Space by not having to solve some components more than once. That is, when they arrive at a node $n_{ao}$ in their generated AND/OR tree $AO$, if $n_{ao}.f$ has been solved before they can immediately backtrack.

#DPLL-Cache gains the same efficiency over #DPLL-Space. In particular, it need never solve the same component more than once. Using caching to removing previously solved components from its list of components gives rise to the same savings that are realized by adding caching to AND/OR or RC. Formally, the same construction of a partial decision tree $DT$ can be used. In $AO$ we mark all nodes where search is terminated by a cache hit as a satisfying node (if the cached formula is satisfiable) or as a failure node (if the cached formula is unsatisfiable). Now, for example, AND nodes can have satisfying or failure nodes as children when those components have been solved before. Applying our construction to $AO$ gives rise to a partial decision tree $DT$, and it can then be shown that #DPLL-Cache using $DT$ to guide its variable choices will explore a search tree that is about the same size as $DT$. This proves that #DPLL-Cache polynomially simulates RC-Cache and AND/OR-Cache.

The only subtle point is that #DPLL-Cache might not solve a component at the same point in its search. In particular, if a component $\phi$ first appears on #DPLL-Cache's list of components with a previously added unsatisfiable component, #DPLL-Cache will backtrack without solving $\phi$. Following $DT$, #DPLL-Cache will only do enough work to find $\phi$'s first solution, after which it will proceed to the other components on its list. During its search for $\phi$'s first solution, it will cache all unsatisfiable reductions of $\phi$ found during this search. Thus, the next time it encounters $\phi$ it can follow the same variable ordering and not do any extra work: the cached unsatisfiable reductions will immediately prune all paths leading to failure and it can proceed directly to the first solution to $\phi$. If the other components on its list are all satisfiable, it will eventually backtrack to this first solution and then continue to solve $\phi$. Hence, although #DPLL-Cache might encounter $\phi$ many times before solving it, each such encounter, except for the first, require adding to its search tree only a number of nodes linear in the number of variables in $\phi$. The number of nodes added by the first encounter, where the $\phi$'s first solution is found, and the encounter where it finally solves $\phi$, together equal the number of nodes required in $AO$ to solve $\phi$. Hence, the "encounters without solving" do not increase the size of #DPLL-Cache's search tree by more than a polynomial.

Finally, we note that the construction given accommodates the use of dynamic variable orderings where the order of variables varies from branch to branch in the AND/OR search tree. (Varying the value assigned along the left and right branch of each query variable is also accommodated). That is, the proof also shows that #DPLL-Cache polynomially simulates AND/OR-Cache$^+$ and RC-Cache$^+$.  □

**Theorem 6** None of RC-Space, RC-Cache, AND/OR-Cache, AND/OR-Space or VE can polynomially simulate #DPLL-Cache, #DPLL-Space, or #DPLL.

To prove this theorem we first observe that from a result of Johannsen (Johannsen, 2001), #DPLL-Cache, #DPLL-Space, and #DPLL can all solve the negation of the propositional string-of-pearls principle (Bonet, Esteban, Galesi, & Johannsen, 1998) in time $n^{O(\log n)}$, when run with a *dynamic* variable ordering. Then we prove (in Theorem 7) that all of the other algorithms require time exponential in $n$ on this problem. Hence, none of these algorithms can polynomially simulate #DPLL (or the stronger #DPLL-Space or #DPLL-Cache).

The string-of-pearls principle, introduced in a different form by Clote and Setzer (1998) and explicitly by Bonet et al. (1998) is as follows. From a bag of $m$ pearls, which are colored red and blue, $n$ pearls are chosen and placed on a string. The string-of-pearls principle says that if the first pearl in the string is red and the last one is blue, then there must be a red-blue or blue-red pair of pearls side-by-side somewhere on the string. The negation of the principle, $S_{m,n}$, is expressed with variables $p_{i,j}$ and $p_j$ for $i \in [n]$ and $j \in [m]$ where $p_{i,j}$ represents whether pearl $j$ is mapped to vertex $i$ on the string, and $p_j$ represents whether pearl $j$ is colored blue ($p_j = 0$) or red ($p_j = 1$). The clauses of $SP_{m,n}$ are as follows.

(1) Each hole gets at least one pearl: $\vee_{j=1}^m p_{i,j}$, $i \in [n]$.

(2) Each hole gets at most one pearl: $(\neg p_{i,j} \vee \neg p_{i,j'})$, $i \in [n]$ $j \in [m]$ ,$j' \in [m]$, $j \neq j'$.

(3) A pearl goes to at most one hole: $(\neg p_{i,j} \vee \neg p_{i',j})$, $i \in [n]$, $i' \in [n]$, $i \neq i'$, $j \in [m]$.

(4) The leftmost hole gets assigned a red pearl and the rightmost hole gets assigned a blue pearl: $(\neg p_{1,j} \vee p_j)$ and $(\neg p_{n,j} \vee \neg p_j)$, $j \in [m]$.

(5) Any two adjacent holes get assigned pearls of the same color: $(\neg p_{i,j} \vee \neg p_{i+1,j'} \vee \neg p_j \vee p_{j'})$, $1 \leq i < n, j \in [m], j' \in [m], j \neq j'$, and $(\neg p_{i,j} \vee \neg p_{i+1,j'} \vee p_j \vee \neg p_{j'})$, $1 \leq i < n, j \in [m]$, $j' \in [m], j \neq j'$.

Johannsen (Johannsen, 2001) shows that $SP_{n,n}$ has quasipolynomial size *tree* resolution proofs. It follows that #DPLL, #DPLL-Space and #DPLL-Cache can solve $SP_{n,n}$ in quasipolynomial time.

**Lemma 4** (Johannsen, 2001) $SP_{n,n}$ can be solved in time $n^{O(\log n)}$ by #DPLL, #DPLL-Space, and #DPLL-Cache.

**Theorem 7** Let $\epsilon = 1/5$. Any of the algorithms RC-Space, RC-Cache, AND/OR-Cache, AND/OR-Space, VE, or #DPLL-Cache using a static variable ordering, require time $2^{n^\epsilon}$ to solve $SP_{n,n}$.

**Proof:** It can be seen from the proof of Theorem 5 that #DPLL-Cache using a static variable ordering can polynomially simulate all of the stated algorithms.

Hence, it suffices to prove that #DPLL-Cache under **any** static ordering requires time $2^{n^\epsilon}$ for $SP_{m,n}$, $m = n$. By a static ordering, we mean that the variables are queried according to this ordering as long as they are mentioned in the current formula. That is, we allow a variable to be skipped over if it is irrelevant to the formula currently under consideration. We will visualize $SP_{n,n}$ as a bipartite graph, with $n$ vertices on the left, and $n$ pearls on the right. There is a pearl variable $p_j$ corresponding to each of the $n$ pearls, and an edge variable $p_{i,j}$ for every vertex-pearl pair. (Note that there are no variables corresponding to the vertices but we will still refer to them.)

Fix a particular total ordering of the underlying $n^2 + n$ variables, $\theta_1, \theta_2, \ldots, \theta_l$. For a pearl $j$, let $fanin_t(j)$ equal the number of edge variables $p_{k,j}$ incident with pearl $j$ that are one of the first $t$ variables queried. Similarly, for a vertex $i$, let $fanin_t(i)$ equal the number of edge variables $p_{i,k}$ incident with vertex $i$ that are one of the first $t$ variables queried. For a set of pearls $S$, let $fanin_t(S)$ equal the number of edge variables $p_{k,j}$ incident with some pearl $j \in S$ that are one of the first $t$ variables queried. Similarly for a set of vertices $S$, $fanin_t(S)$ equals the number of edge variables $p_{i,k}$ incident with some vertex $i \in S$ that are one of the first $t$ variables queried. Let $edges_t(j)$ and $edges_t(S)$ be defined similarly although now it is the set of such edges rather than the number of such edges. It should be clear from the context whether the domain objects are pearls or vertices.

We use a simple procedure, based on the particular ordering of the variables, for marking each pearl with either a **C** or with an **F** as follows. In this procedure, a pearl may at some point be marked with a **C** and then later overwritten with an **F**; however, once a pearl is marked with an **F**, it remains an **F** for the duration of the procedure. If a pearl $j$ is marked with a **C** at some particular point in time, $t$, this means that at this point, the color of the pearl has already been queried, and $fanin_t(j)$ is less than $n^\delta$, $\delta = 2/5$. If a pearl $j$ is marked with an **F** at some particular point in time $t$, it means that at this point $fanin_t(j)$ is at least $n^\delta$. (The color of $j$ may or may not have been queried.) If a pearl $j$ is unmarked at time $t$, this means that its color has not yet been queried, and $fanin_t(j)$ is less than $n^\delta$.

For $l$ from 1 to $n^2+n$, we do the following. If the $l^{th}$ variable queried is a pearl variable ($\theta_l = p_j$ for some $j$), and less than $n^\delta$ edges $p_{i,j}$ incident to $j$ have been queried so far, then mark $p_j$ with a **C**. Otherwise, if the $l^{th}$ variable queried is an edge variable ($\theta_l = p_{i,j}$) and $fanin_l(j) \geq n^\delta$, then mark pearl $j$ with an **F** (if not already marked with an **F**). Otherwise, leave pearl $j$ unmarked.

Eventually every pearl will become marked **F**. Consider the first time $t^*$ where we have either a lot of **C**'s, or a lot of **F**'s. More precisely, let $t^*$ be the first time where either there are exactly $n^\epsilon$ **C**'s (and less than this many **F**'s) or where there are exactly $n^\epsilon$ **F**'s (and less than this many **C**'s.) If exactly $n^\epsilon$ **C**'s occurs first, then we will call this case (a). Extend $t^*$ to $t_a^*$ as follows. Let $\theta_{t^*+1}, \ldots, \theta_{t^*+c}$ be the largest segment of variables that are all pearl variables $p_j$ such that $j$ is already marked with an **F**. Then $t_a^* = t^* + c$. Notice that the query immediately following $\theta_{t_a^*}$ is either a pearl variable $p_j$ that is currently unmarked, or an edge variable. On the other hand, if exactly $n^\epsilon$ **F**'s occurs first, then we will call this case (b). Again, extend $t^*$ to $t_b^*$ to ensure that the query immediately following $\theta_{t_b^*}$ is either a pearl variable $p_j$ that is currently unmarked, or is an edge variable.

The intuition is that in case (a) (a lot of **C**'s), a lot of pearls are colored prematurely–that is, before we know what position they are mapped to–and hence a lot of queries must be asked. For case (b) (a lot of **F**'s), a lot of edge variables are queried thus again a lot of queries will be asked. We now proceed to prove this formally.

We begin with some notation and definitions. Let $f = SP_{n,n}$, and let $Vars(f)$ denote the set of all variables underlying $f$. A restriction $\rho$ is a partial assignment of some of the variables underlying $f$ to either 0 or to 1. If a variable $x$ is unassigned by $\rho$, we denote this by $\rho(x) = *$. Let $T$ be the DPLL tree based on the variable ordering $\theta$. That is, $T$ is a decision tree where variable $\theta_i$ is queried at level $i$ of $T$. Recall that corresponding to each node $v$ of $T$ is a formula $f|_\rho$ where $\rho$ is the restriction corresponding to the partial path from the root of $T$ to $v$. The tree $T$ is traversed by a depth-first search. For each vertex $v$ with corresponding path $p$ that is traversed, we check to see if $f|_p$ is already in the cache. If it is, then there is no need to traverse the subtree rooted below $v$. If it is not yet in the cache, then we traverse the left subtree of $v$, followed by the right subtree of $v$. After both subtrees have been traversed, we then pop back up to $v$, and store $f|_p$ in the cache. This induces an ordering on the vertices (and corresponding paths) of $T$ that are traversed—whenever we pop back up to a vertex $v$ (and thus, we can store its value in the cache), we put $v$ ($p$) at the end of the current order.

**Lemma 5** Let $f$ be $SP_{n,n}$ and let $\pi$ be a static ordering of the variables. Let $\rho$ be a partial restriction of the variables. Then the runtime of #DPLL-Cache on $(f, \rho)$ is not less than the runtime of #DPLL-Cache on $(f|_\rho, \pi')$, where $\pi'$ is the ordering of the unassigned variables consistent with $\pi$.

**Lemma 6** For any restriction $\rho$, if $f|_\rho \neq 0$ and $\rho(p_{i,j}) = *$, then $p_{i,j}$ occurs in $f|_\rho$.

**Proof:** Consider the clause $C_i = (p_{i,1} \vee \ldots \vee p_{i,m})$ in $f$. Since $p_{i,j}$ is in this clause, if $p_{i,j}$ does not occur in $f|_\rho$, then $C_i|_\rho$ must equal 1. Thus there exists $j' \neq j$ such that $\rho(p_{i,j'}) = 1$. But then the clause $(\neg p_{i,j} \vee \neg p_{i,j'})|_\rho = \neg p_{i,j}$ and thus $p_{i,j}$ does not disappear from $f|_\rho$. $\quad\square$

**Corollary 1** Let $\theta$ be a total ordering of $Vars(f)$. Let $\rho$, $\rho'$ be partial restrictions such that $\rho$ sets exactly $\theta_1, \ldots, \theta_q$ and $\rho'$ sets exactly $\theta_1, \ldots, \theta_{q'}$, $q' < q$. Suppose that there exists $\theta_k = p_{i,j}$ such that $\rho$ sets $\theta_k$ but $\rho'(\theta_k) = *$. Then either $f|_\rho = 0$ or $f|_{\rho'} = 0$ or $f|_\rho \neq f|_{\rho'}$.

**Case (a).** Let $\theta$ be a total ordering to $Vars(f)$ such that case (a) holds. Let $P^C$ denote the set of exactly $n^\epsilon$ pearls that are marked **C** and let $P^F$ denote the set of less than $n^\epsilon$ pearls (disjoint from $P^C$) that are marked **F**. Note that (the color of) all pearls in $P^C$ have been queried by time $t_a^*$; the color of the pearls in $P^F$ may be queried by time $t_a^*$, and the color of all pearls in $P - P^C - P^F$ have not been queried by time $t_a^*$. Note further that the total number of edges $p_{i,j}$ that have been queried is at most $n^{\epsilon+\delta} + n^{1+\epsilon} \leq 2n^{1+\epsilon}$.

We will define a partial restriction, $M_a$, to all but $2^{n^\epsilon}$ of the variables in $\theta_1, \ldots, \theta_{t_a^*}$ as follows. For each $j \in P^F$, fix a one-to-one mapping from $P^F$ to $[n]$ such that $range(j) \in edges_{t_a^*}(j)$ for each $j$. For each $j \in P^C$, for any variable $p_{i,j}$ queried in $\theta_1, \ldots \theta_{t_a^*}$, set $p_{i,j}$ to 0. For any vertex $i$ such that all variables $p_{i,j}$ have been queried in $\theta_1, \ldots, \theta_{t_a^*}$, map $i$ to exactly one pearl $j$ such that $p_j \in P - P^C - P^F$. There are at most $2n^\epsilon$ such $i$. (This can be arbitrary as long as it is consistent with the one-to-one mapping already defined on $P^F$.) For all remaining $p_j \in P - P^C - P^F$ that have not yet been mapped to, set all queried variables $p_{i,j}$ to 0. For all pearls $p_{i,j}$ in $P^F$ that have been queried in $\theta_1, \ldots, \theta_{t_a^*}$, assign a fixed color to each such pearl (all Red or all Blue) so that the smallest Red/Blue gap is as large as possible. Note that the gap will be of size at least $n^{1-\epsilon}$. $M_a$ sets all variables in $\theta_1, \ldots \theta_{t_a^*}$ except for the variables $p_j$, $j \in P^C$. Since there are $n^\epsilon$ such variables, the number of restrictions $\rho$ to $\theta_1, \ldots, \theta_{t_a^*}$ consistent with $M_a$ is exactly $2^{n^\epsilon}$. Let $S$ denote this set of restrictions.

Let $f' = f|_{M_a}$ and let $\theta'$ be the ordering on the unassigned variables consistent with $\theta$. (The set of unassigned variables is: $p_j$, for $j \in P^C$, plus all variables in $\theta_k$, $k > t_a^*$.) Let $T'$ be the DPLL tree corresponding to $\theta'$ for solving $f'$. By Lemma 5, it suffices to show that #DPLL-Cache when run on inputs $f'$ and $T'$, takes time at least $2^{n^\epsilon}$.

Note that the first $n^\epsilon$ variables queried in $T'$ are the pearl variables in $P^C$, and thus the set of all $2^{n^\epsilon}$ paths of height exactly $n^\epsilon$ in $T'$ correspond to the set $S$ of all possible settings to these variables. We want to show that for each vertex v of height $n^\epsilon$ in $T'$ (corresponding to each of the $2^{n^\epsilon}$ settings of all variables in $P^C$), that $v$ must be traversed by #DPLL-Cache, and thus the runtime is at least $2^{n^\epsilon}$.

Fix such a vertex $v$, and corresponding path $\rho \in S$. If $v$ is not traversed, then there is some $\rho' \subseteq \rho$ and some $\sigma$ such that $\sigma$ occurs before $\rho'$ in the ordering, and such that $f'|_\sigma = f'|_{\rho'}$. We want to show that this cannot happen. There are several cases to consider.

**1a.** Suppose that $|\sigma| \leq n^\epsilon$ and $\sigma \neq \rho'$. Then both $\rho'$ and $\sigma$ are partial assignments to some of the variables in $P^C$ that are inconsistent with one another. It is easy to check that in this case, $f'|_{\rho'} \neq f'|_\sigma$.

**2a.** Suppose that $|\sigma| > n^\epsilon$, and the $(n^\epsilon + 1)^{st}$ variable set by $\sigma$ is an edge variable $p_{i,j}$. Because $|\rho'| \leq n^\epsilon$, $\rho'(p_{i,j}) = *$. By Corollary 1, it follows that $f'|_{\rho'} \neq f'|_\sigma$.

**3a.** Suppose that $|\sigma| > n^\epsilon$ and the $(n^\epsilon + 1)^{st}$ variable set by $\sigma$ is a pearl variable $p_j$. (Again, we know that $p_j$ is unset by $\rho'$.) Since this is case (a), we can assume that $p_j \in P - P^C - P^F$. Call a vertex $i$ *bad* if $P - P^F - P^C \subset edges_{t_a^*}(i)$. If $i$ is bad, then $fanin_{t_a^*}(i)$ is greater than $n - 2n^\epsilon \geq n/2$. Since the total number of edges queried is at most $2n^{1+\epsilon}$, if follows that the number of bad vertices is at most $4n^\epsilon$. This implies that we can find a pair $i, i + 1$ of vertices and a pearl $j'$ such that: (1) $p_{i,j}$ is not queried in $\theta_1, \ldots, \theta_{t_a^*}$; (2) $p_{i+1,j'}$ is not queried in $\theta_1, \ldots, \theta_{t_a^*}$; (3) $p_{j'}$ is in $P - P^C - P^F$ and thus $p_{j'}$ is also not queried. Thus the clause $(\neg p_{i,j} \vee \neg p_j \vee \neg p_{i+1,j'} \vee p_{j'})|_{\rho'}$ does not disappear or shrink in $f'|_{\rho'}$, and thus $f'|_{\rho'} \neq f'|_\sigma$.

**Case (b).** Let $\theta$ be a total ordering to $Vars(f)$ such that case (b) holds. Now let $P^C$ denote the set of less than $n^\epsilon$ pearls marked **C** and let $P^F$ denote the set of exactly $n^\epsilon$ pearls marked **F**.

We define a partial restriction $M_b$ to all but $2^{n^\epsilon}$ of the variables in $\theta_1, \ldots, \theta_{t^*}$ as follows. Call a vertex $i$ *full* if all variables $p_{i,j}$ have been queried in $\theta_1, \ldots, \theta_{t_b^*}$. There are at most $n^\epsilon$ full vertices. For each $j \in P^F$, we will fix a pair of vertices $F_j = (i_j, i'_j)$ in $[n]$. Let the union of all $n^\epsilon$ sets $F_j$ be denoted by $F$. $F$ has the following properties. (1) For each $j$, no element of $F_j$ is full; (2) For each $j \in P^F$, $F_j \in edges_{t_b^*}(j)$; and (3) every two distinct elements in $F$ are at least distance 4 apart. Since $fanin_{t_b^*}(j) \geq n^\delta$, and $\delta = 2/5 > \epsilon$, it is possible to find such sets $F_j$ satisfying these criteria.

For each $p_{i,j}$ queried in $\theta_1, \ldots, \theta_{t_b^*}$, where $j \in P^F$ and $i \notin F_j$, $M_b$ will set $p_{i,j}$ to 0. For each $j \in P^C$, and for any variable $p_{i,j}$ queried in $\theta_1, \ldots, \theta_{t_b^*}$, set $p_{i,j}$ to 0. For any full vertex $i$, map $i$ to exactly one pearl $j$ such that $p_j \in P - P^C - P^F$. (Again this can be arbitrary as long as it is consistent with a one-to-one mapping.) For the remaining $p_j \in P - P^C - P^F$ that have not yet been mapped to, set all queried variables $p_{i,j}$ to 0. For all pearls $p_j$ in $P^C$, color them Red. For all pearls $p_j$ in $P^F$ that have been queried, assign a fixed color to each pearl.

The only variables that were queried in $\theta_1, \ldots, \theta_{t_b^*}$ and that are not set by $M_b$ are the edge variables, $p_{i,j}$, where $j \in P^F$, and $i \in F_j$. Let $S$ denote the set of all $2^{n^\epsilon}$ settings of these edge variables such that each $j \in P^F$ is mapped to exactly one element in $F_j$. Let $f' = f|_{M_b}$ and let $T'$ be the DPLL tree corresponding to $\theta'$ for solving $f'$, where $\theta'$ is the ordering on the unassigned variables consistent with $\theta$. By Lemma 5, it suffices to show that #DPLL-Cache on $f'$ and $T'$ takes time at least $2^{n^\epsilon}$.

Note that the first $2n^\epsilon$ variables queried in $T'$ are the variables $P_{i,j}$, $P_{i'_j,j}$, $j \in P^F$. The only nontrivial paths of height $2n^\epsilon$ in $T'$ are those were each $j \in P^F$ is mapped to exactly one vertex in $F_j$, since otherwise the formula $f'$ is set to 0. Thus, the nontrivial paths in $T'$ of height $2n^\epsilon$ correspond to $S$. We want to show that for each such nontrivial vertex $v$ of height $2n^\epsilon$ in $T'$ (corresponding to each of the restrictions in $S$), that $v$ must be traversed by #DPLL-Cache, and thus the runtime is at least $2^{n^\epsilon}$.

Fix a vertex $v$ and corresponding path $\rho \in S$. Again we want to show that for any $\rho' \subseteq \rho$, and $\sigma$ where $\sigma$ occurs before $\rho'$ in the ordering, that $f'|_{\rho'} \neq f'|_\sigma$. There are three cases to consider.

**1b.** Suppose that $|\sigma| \leq 2n^\epsilon$. If $\sigma$ is nontrivial, then both $\rho'$ and $\sigma$ are partial mappings of the pearls $j$ in $P^F$ to $F_j$, that are inconsistent with one another. It is easy to check that in this case $f'|_\sigma \neq f'|_{\rho'}$.

**2b.** Suppose that $|\sigma| > 2n^\epsilon$ and the $(2n^\epsilon + 1)^{st}$ variable set by $\sigma$ is an edge variable $p_{i,j}$. Because $|\rho'| \leq 2n^\epsilon$, $\rho'(p_{i,j}) = *$. By Corollary 1, it follows that $f'|_\sigma \neq f'|_{\rho'}$.

**3b.** Suppose that $|\sigma| > 2n^\epsilon$ and the $(2n^\epsilon + 1)^{st}$ variable set by $\sigma$ is a pearl variable $p_j$. By the definition of $t_b^*$, we can assume that $p_j \in P - P^C - P^F$. By reasoning similar to case 3a, can find vertices $i, i+1$, and pearl $j' \in P - P^C - P^F$ such that none of the variable $p_{i,j}, p_{i+1,j}, p_{j'}$ are queried in $\theta_1, \ldots, \theta_{t_b^*}$. Thus the clause $(\neg p_{i,j} \vee \neg p_j \vee \neg p_{i+1,j'} \vee p_{j'})|_{\rho'}$ does not disappear to shrink in $f'|_{\rho'}1$, and therefore $f'|_{\rho'} \neq f'|_\sigma$.

Thus for each of the two cases, #DPLL-Cache on $f'$ and $T'$ takes time at least $2^{n^\epsilon}$ and thus #DPLL-Cache on $f$ and $T$ takes time at least $2^{n^\epsilon}$. $\quad \square$

# References

Aleknovich, A., & Razborov, A. (2002). Satisfiability, Branch-width and Tseitin Tautologies. In *Annual IEEE Symposium on Foundations of Computer Science (FOCS)*, pp. 593–603.

Bacchus, F., Dalmao, S., & Pitassi, T. (2003). Algorithms and Complexity Results for #SAT and Bayesian Inference. In *Annual IEEE Symposium on Foundations of Computer Science (FOCS)*, pp. 340–351.

Bayardo, R. J., & Pehoushek, J. D. (2000). Counting Models using Connected Components. In *Proceedings of the AAAI National Conference (AAAI)*, pp. 157–162.

Bayardo, R. J., & Miranker, D. P. (1995). On the space-time trade-off in solving Constraint Satisfaction Problems. In *Proceedings of the International Joint Conference on Artificial Intelligence (IJCAI)*, pp. 558–562.

Beame, P., Impagliazzo, R., Pitassi, T., & Segerlind, N. (2003). Memoization and DPLL: Formula Caching Proof Systems. In *IEEE Conference on Computational Complexity*, pp. 248–264.

Birnbaum, E., & Lozinskii, E. L. (1999). The good old Davis Putnam procedure helps counting models. *J. Artif. Intell. Research (JAIR)*, *10*, 457–477.

Bitner, J. R., & Reingold, E. (1975). Backtracking programming techniques. *Communications of the ACM*, *18*(11), 651–656.

Bodlaender, H. L. (1993). A tourist guide through Treewidth. *Acta Cybernetica*, *11*(1–2), 1–21.

Bonet, M., Esteban, J. L., Galesi, N., & Johannsen, J. (1998). Exponential separations between restricted resolution and cutting planes proof systems. In *Annual IEEE Symposium on Foundations of Computer Science (FOCS)*, pp. 638–647.

Boutilier, C., Friedman, N., Goldszmidt, M., & Koller, D. (1996). Context-specific independence in Bayesian Networks. In *Uncertainty in Artificial Intelligence, Proceedings of Annual Conference (UAI)*, pp. 115–123.

Chavira, M., & Darwiche, A. (2006). Encoding CNFs to empower component analysis. In *Theory and Applications of Satisfiability Testing (SAT)*, pp. 61–74.

Chavira, M., & Darwiche, A. (2008). On probabilistic inference by weighted model counting. *Artificial Intelligence*, *172*(6-7), 772–799.

Chavira, M., Darwiche, A., & Jaeger, M. (2006). Compiling relational bayesian networks for exact inference. *Int. J. Approx. Reasoning*, *42*(1-2), 4–20.

Clote, P., & Setzer, A. (1998). On PHP, st-connectivity and odd charged graphs. In *Proof Complexity and Feasible Arithmetics*, Vol. 39 of *DIMACS Series*, pp. 93–117. AMS.

Cormen, T. H., Leiserson, C. E., Rivest, R. L., & Stein, C. (2001). *Introduction to Algorithms. 2nd Edition*. McGraw Hill.

Darwiche, A., & Allen, D. (2002). Optimal time-space tradeoff in probabilistic inference. In *European Workshop on Probabilistic Graphical Models*. Available at www.cs.ucla.edu/~darwiche.

Darwiche, A. (2001). Recursive conditioning. *Artificial Intelligence*, *126*, 5–41.

Darwiche, A. (2002). A logical approach to factoring belief networks. In *Proceedings of the International Conference on Principles of Knowledge Representation and Reasoning*, pp. 409–420.

Darwiche, A. (2004). New advances in compiling CNF into decomposable negation normal form. In *Proceedings of the European Conference on Artificial Intelligence (ECAI)*, pp. 328–332.

Davies, J., & Bacchus, F. (2007). Using more reasoning to improve #SAT solving. In *Proceedings of the AAAI National Conference (AAAI)*, pp. 185–190.

Davis, M., Logemann, G., & Loveland, D. (1962). A machine program for theorem-proving. *Communications of the ACM*, *4*, 394–397.

Davis, M., & Putnam, H. (1960). A computing procedure for quantification theory. *Journal of the ACM*, *7*, 201–215.

Dechter, R. (1999). Bucket elimination: A unifying framework for reasoning. *Artificial Intelligence*, *113*, 41–85.

Dechter, R., & Mateescu, R. (2004). Mixtures of deterministic-probabilistic networks and their AND/OR search space. In *Uncertainty in Artificial Intelligence, Proceedings of Annual Conference (UAI)*, pp. 120–129.

Dechter, R., & Mateescu, R. (2007). AND/OR search spaces for graphical models. *Artificial Intelligence*, *171*(2-3), 73–106.

Dubois, O. (1991). Counting the number of solutions for instances of satisfiability. *Theoretical Computer Science*, *81*, 49–64.

Haken, A. (1985). The intractability of resolution. *Theoretical Computer Science*, *39*, 297–305.

Hertel, P., Bacchus, F., Pitassi, T., & van Gelder, A. (2008). Clause learning can effectively p-simulate general propositional resolution. In *Proceedings of the AAAI National Conference (AAAI)*.

Johannsen, J. (2001). Exponential incomparability of tree-like and ordered resolution. Unpublished manuscript, available at `http://www.tcs.informatik.uni-muenchen.de/~jjohanns/notes.html`.

Kask, K., Dechter, R., Larrosa, J., & Dechter, A. (2005). Unifying tree decompositions for reasoning in graphical models. *Artificial Intelligence*, *166*(1-2), 165–193.

Kitching, M., & Bacchus, F. (2008). Exploiting decomposition in constraint optimization problems. In *Proceedings of Principles and Practice of Constraint Programming (CP)*, pp. 478–492.

Lauritzen, S., & Spiegelhalter, D. (1988). Local computation with probabilities on graphical structures and their application to expert systems. *Journal of the Royal Statistical Society Series B*, *50*(2), 157–224.

Li, W., & van Beek, P. (2004). Guiding real-world sat solving with dynamic hypergraph separator decomposition. In *Proceedings of the International Conference on Tools with Artificial Intelligence (ICTAI)*, pp. 542–548.

Li, W., van Beek, P., & Poupart, P. (2006). Performing incremental Bayesian Inference by dynamic model counting. In *Proceedings of the AAAI National Conference (AAAI)*, pp. 1173–1179.

Li, W., van Beek, P., & Poupart, P. (2008). Exploiting causal independence using weighted model counting. In *Proceedings of the AAAI National Conference (AAAI)*.

Littman, M. L., Majercik, S. M., & Pitassi, T. (2001). Stochastic boolean satisfiability. *J. Automated Reasoning*, *27*(3), 251–296.

Majercik, S. M., & Littman, M. L. (1998). Maxplan: A new approach to probabilistic planning. In *Proceedings of the International Conference on Artificial Intelligence Planning and Scheduling (AIPS)*, pp. 86–93.

Marinescu, R., & Dechter, R. (2006). Dynamic orderings for AND/OR branch-and-bound search in graphical models. In *Proceedings of the European Conference on Artificial Intelligence (ECAI)*, pp. 138–142.

Marinescu, R., & Dechter, R. (2007). Best-first AND/OR search for graphical models. In *Proceedings of the AAAI National Conference (AAAI)*, pp. 1171–1176.

Mateescu, R., & Dechter, R. (2007). AND/OR multi-valued decision diagrams for weighted graphical models. In *Uncertainty in Artificial Intelligence, Proceedings of Annual Conference (UAI)*.

Mateescu, R., & Dechter, R. (2005). AND/OR cutset conditioning. In *Proceedings of the International Joint Conference on Artificial Intelligence (IJCAI)*, pp. 230–235.

Moskewicz, E., Madigan, C., Zhao, M., Zhang, L., & Malik, S. (2001). Chaff: Engineering an efficient sat solver. In *Proc. of the Design Automation Conference (DAC)*.

Nilsson, N. J. (1980). *Principles of Artificial Intelligence*. Tioga.

Pearl, J. (1988). *Probabilistic Reasoning in Intelligent Systems* (2nd edition). Morgan Kaufmann, San Mateo, CA.

Preston, C. (1974). *Gibbs States on Countable Sets*. Cambridge University Press.

Rish, I., & Dechter, R. (2000). Resolution versus search: Two strategies for SAT. *Journal of Automated Reasoning*, *24*(1), 225–275.

Robertson, N., & Seymour, P. (1991). Graph minors X. obstructions to tree-decomposition. *Journal of Combinatorial Theory, Series B*, *52*, 153–190.

Robertson, N., & Seymour, P. (1995). Graph minors XIII. the disjoint paths problem. *Journal of Combinatorial Theory, Series B*, *63*, 65–110.

Roth, D. (1996). On the hardness of approximate reasoning. *Artificial Intelligence*, *82*(1–2), 273–302.

Sang, T., Bacchus, F., Beame, P., Kautz, H. A., & Pitassi, T. (2004). Combining component caching and clause learning for effective model counting. In *Theory and Applications of Satisfiability Testing (SAT)*.

Sang, T., Beame, P., & Kautz, H. A. (2005a). Heuristics for fast exact model counting. In *Theory and Applications of Satisfiability Testing (SAT)*, pp. 226–240.

Sang, T., Beame, P., & Kautz, H. A. (2005b). Performing Bayesian Inference by weighted model counting. In *Proceedings of the AAAI National Conference (AAAI)*, pp. 475–482.

Sang, T., Beame, P., & Kautz, H. A. (2007). A dynamic approach for MPE and weighted MAX-SAT. In *Proceedings of the International Joint Conference on Artificial Intelligence (IJCAI)*, pp. 173–179.

Sanner, P., & McAllester, D. (2005). Affine algebraic decision diagrams (aadds) and their applications to structured probabilistic inference. In *Proceedings of the International Joint Conference on Artificial Intelligence (IJCAI)*, pp. 1384–1390.

Spitzer, F. L. (1971). Markov random fields and Gibbs ensembles. *American Mathematical Monthly*, *78*, 142–54.

Thurley, M. (2006). sharpSAT—Counting models with advanced component caching and implicit BCP. In *Theory and Applications of Satisfiability Testing (SAT)*, pp. 424–429.

Valiant, L. G. (1979a). The complexity of enumeration and reliability problems. *SIAM Journal of Computing*, *9*, 410–421.

Valiant, L. G. (1979b). The Complexity of Computing the Permanent. *Theoretical Computer Science*, *8*, 189–201.

Zhang, W. (1996). Number of models and satisfiability of sets of clauses. *Theoretical Computer Science*, *155*, 277–288.